\pdfoutput=1

\documentclass[11pt]{article}

\usepackage[final]{acl}

\usepackage{times}
\usepackage{latexsym}

\usepackage[T1]{fontenc}

\usepackage[utf8]{inputenc}

\usepackage{microtype}

\usepackage{inconsolata}

\usepackage{graphicx}

\usepackage{pifont}
\usepackage{hyperref}
\usepackage{amssymb}
\usepackage{caption}
\usepackage{makecell}
\usepackage{multirow}
\usepackage{float}
\usepackage{enumitem}
\usepackage{array}
\usepackage{amsmath}
\usepackage[capitalise]{cleveref}
\usepackage{listings}
\usepackage{longtable}
\usepackage{booktabs}
\usepackage{xcolor}
\usepackage{subcaption}
\usepackage[framemethod=TikZ]{mdframed}
\usepackage{etoolbox}
\usepackage[listings]{tcolorbox}
\tcbuselibrary{listings,theorems}

\definecolor{UserExampleBg}{HTML}{FAFDF9}
\newmdenv[
    roundcorner=5pt,
    backgroundcolor=UserExampleBg,
    linecolor=black,
    outerlinewidth=1pt,
    frametitlebackgroundcolor=black,
    frametitlefont={\bfseries\color{white}},
]{user_example}
\AtBeginEnvironment{user_example}{\setlength{\parindent}{0pt}}
\newtcbtheorem[]{prompt}{Prompt}%
{colback=green!5,colframe=black,fonttitle=\bfseries, left=.02in, right=.02in,bottom=.02in, top=.02in}{prompt}

\definecolor{deepblue}{rgb}{0,0,0.5}
\definecolor{deepred}{rgb}{0.6,0,0}
\definecolor{deepgreen}{rgb}{0,0.5,0}
\definecolor{pink}{RGB}{219, 48, 122}
\definecolor{forestgreen}{RGB}{34,139,34}
\definecolor{goldenrod}{RGB}{218,165,32}
\definecolor{sepia}{RGB}{112,66,20}
\renewcommand\thanks[1]{\footnotetext[0]{#1}} 

\title{SusGen-GPT: A Data-Centric LLM for Financial NLP and Sustainability Report Generation}

\author{
    Qilong Wu\textsuperscript{\rm 1}\footnotemark[2], 
    Xiaoneng Xiang\textsuperscript{\rm 1}, 
    Hejia Huang\textsuperscript{\rm 1}, 
    Xuan Wang\textsuperscript{\rm 1}, 
    Yeo Wei Jie\textsuperscript{\rm 2}, \\
    {\bf Ranjan Satapathy}\textsuperscript{\rm 3},
    {\bf Ricardo Shirota Filho}\textsuperscript{\rm 3}, 
    {\bf Bharadwaj Veeravalli}\textsuperscript{\rm 1}\\
    \textsuperscript{\rm 1} National University of Singapore, Singapore \\
    \textsuperscript{\rm 2} Nanyang Technological University, Singapore \\
    \textsuperscript{\rm 3} Institute of High Performance Computing (IHPC),\\ Agency for Science, Technology and Research (A* STAR), Singapore \\
    \small Code: \href{https://github.com/JerryWu-code/SusGen}{https://github.com/JerryWu-code/SusGen}
}

\begin{document}
\maketitle
\setcounter{footnote}{0} 
\renewcommand{\thefootnote}{\dag} 
\footnotetext[2]{Correspondence to Qilong Wu: \href{mailto: qilong_wu@u.nus.edu}{qilong\_wu@u.nus.edu}.}
\renewcommand{\thefootnote}{\arabic{footnote}} 

\begin{abstract}
The rapid growth of the financial sector and the increasing focus on Environmental, Social, and Governance (ESG) considerations have created a pressing need for advanced natural language processing (NLP) tools. Despite recent advancements, there is still a notable absence of open-source Large Language Models (LLMs) that are proficient across both general finance and ESG domains, such as generating ESG reports. To address this gap, we introduce \textit{SusGen-30K}, a high-quality, category-balanced dataset comprising seven financial NLP tasks and ESG report generation. In addition, we propose \textit{TCFD-Bench}, a benchmark designed to improve the evaluation of sustainability report generation. Our data-centric approach led to the development of a suite of models, \textit{SusGen-GPT}, trained on the curated dataset. These models were evaluated across six adapted tasks and two off-the-shelf tasks, showing state-of-the-art performance, surpassing all other models except GPT-4. Remarkably, SusGen-GPT achieved an average score only 0.02 below GPT-4, despite using models with only 7-8B parameters compared to GPT-4’s 1,700B. This demonstrates the efficiency of our approach in delivering high performance with significantly fewer resources, addressing existing challenges and fostering further advancements in the financial and ESG research community.

\end{abstract}

\section{Introduction}

As automation in the financial sector gains more attention and climate change becomes an increasingly urgent issue, there is a growing need for more advanced tools capable of solving these problems. The ability to efficiently process and generate financial reports, analyze ESG metrics, and produce comprehensive TCFD-format\footnotemark{} reports is crucial for maintaining transparency and accountability. Recently, LLMs \citep{brown2020language,ouyang2022training,touvron2023llama,gpt4techreport,touvron2023llama2} have gained widespread popularity due to their ability to solve complex tasks such as commonsense reasoning, machine translation or even performing self-training~\citep{yeo2024self}. However, there remains a substantial gap in LLMs tailored towards specialized domains such as finance and ESG \citep{2023finnlp, wang2023fingptbenchmark}. This is likely due to the training dataset being exposed to mostly general text documents and lack data of specialized information.
\footnotetext{\href{https://www.fsb-tcfd.org/}{https://www.fsb-tcfd.org/}. We investigated various sustainability reporting guidelines, including GRI, SASB, EU CSRD, etc., and ultimately chose TCFD because it is more standardized and universally applicable.}

\begin{figure}[ht]
    \centering
    \begin{subfigure}[b]{0.95\linewidth} 
        \centering
        \includegraphics[width=\linewidth]{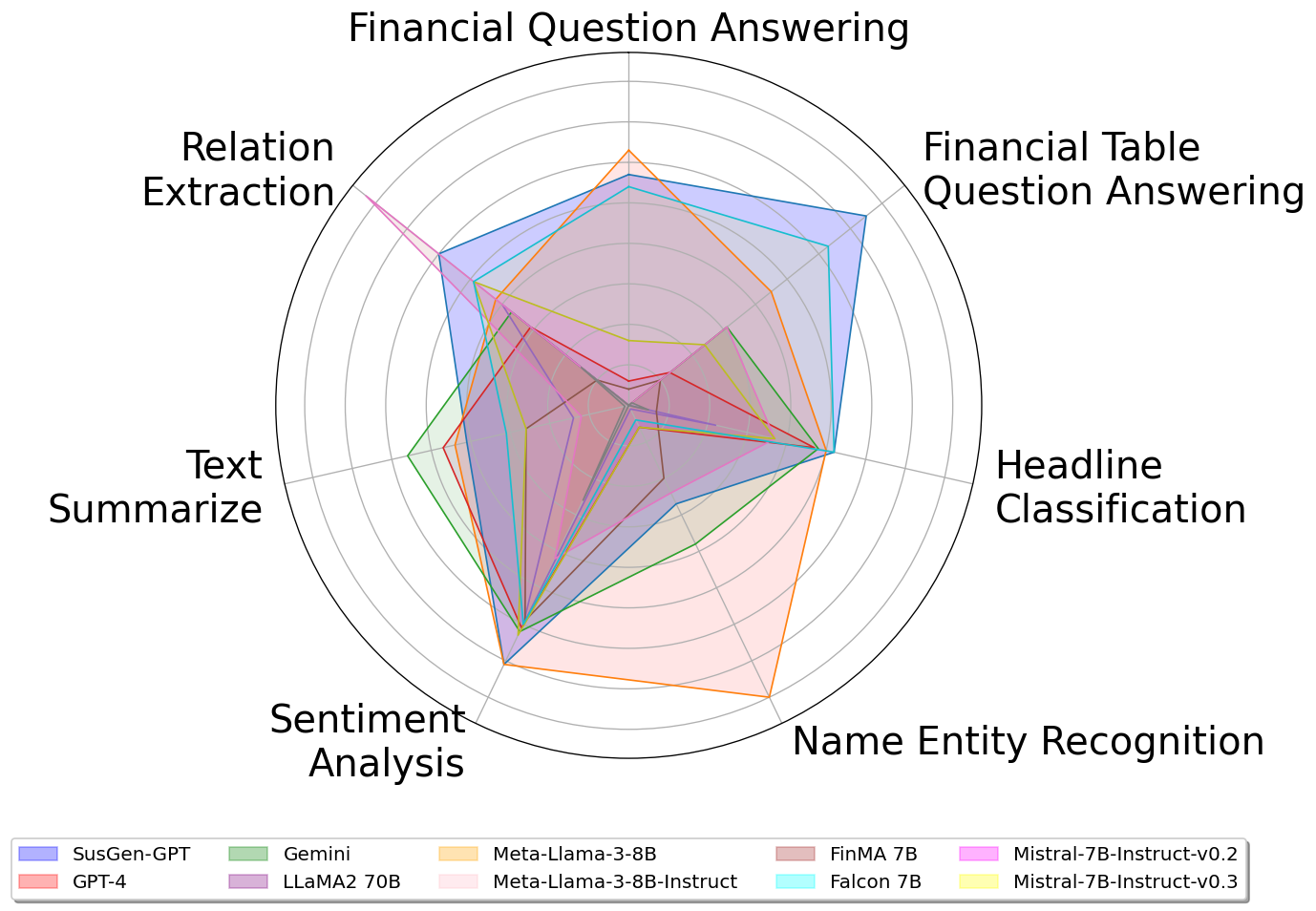}
        \caption{Performance comparison (Relative) with baseline models on each financial NLP task with average score.}
        \label{fig:result_radar}
    \end{subfigure}
    \vfill
    \begin{subfigure}[b]{0.95\linewidth} 
        \centering
        \includegraphics[width=\linewidth]{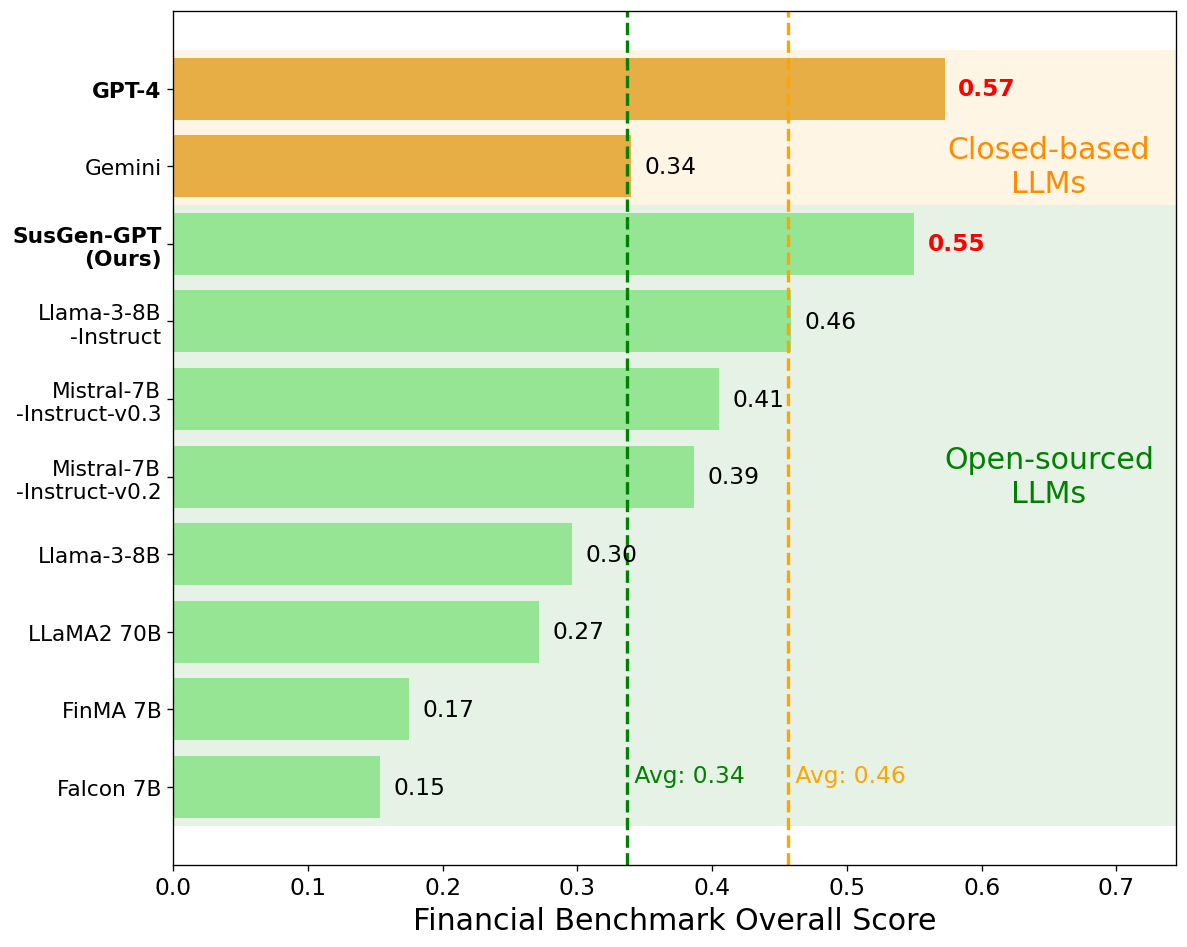}
        \caption{Overall scores of financial NLP benchmark across 6 tasks. Dashed lines for two LLM types’ average.}
        \label{fig:result_bar}
    \end{subfigure}
    \caption{An overview of model comparison with both open-source and closed-base baseline models on six financial NLP tasks. The two sub-figures show that our model SusGen-GPT achieves state-of-the-art performance in most benchmarks.}
    \label{fig:result_merge}
\end{figure}


In light of these challenges, we propose SusGen-30K, a meticulously curated dataset that is designed to tackle multiple NLP tasks across both financial and ESG domains. More importantly, we introduce a suite of LLMs, trained on our proposed dataset, which we refer to as SusGen-GPT. SusGen-GPT is capable of achieving superior performance across multiple downstream tasks simultaneously, when compared against models that are orders of magnitude larger.

Additionally, we propose a new benchmark, TCFD-Bench, specifically designed to assess models’ ability to generate concise and accurate ESG reports from annual reports. We likewise conduct experiments on the proposed benchmark using SusGen-GPT.
In total, our contributions include the following:
\begin{enumerate}[itemsep=0pt,topsep=1pt]
    \item SusGen-30K, a large-scale high quality dataset in both financial \& ESG domain.
    \item We propose and release a well-curated benchmark, tailored towards ESG report generation, TCFD-Bench.
    \item A suite of fine-tuned LLMs, SusGen-GPT, shown to achieve comparable state-of-the-art performance to GPT-4 across both general financial and ESG NLP benchmarks as shown in Figure \ref{fig:result_merge} when most of open-source models struggle to perform well in these domains. Remarkably, our models only have only 7-8B parameter models even with quantization comparing to the GPT4's 1,700B parameters or so, making it a computational efficient solution.
\end{enumerate}

\section{Related Work and Background}

\paragraph{NLP for Finance \& ESG}
Natural Language Processing (NLP) has found extensive applications in various financial tasks, demonstrating its versatility and depth in addressing diverse financial issues \citep{masson-paroubek-2024-evaluating, aguda-etal-2024-large}. The key tasks in the financial domain include Question Answering (QA), Headline Classification (HC), and report generation. More notably, there exists a gap in achieving an acceptable level of proficiency in automating the generation of ESG reports. One such effort, ChatReport \citep{ni-etal-2023-chatreport} is developed to perform summarization and analysis on ESG reports. However, these tools face challenges such as generating reports that are overly simplified and lacking important details. Other attempts~\citep{bronzini2024glitter, zou2023esgreveal, luccioni2020analyzing} mainly rely on existing data extraction techniques and face difficulty in processing unstructured data. Our dataset aims to bridge these gaps by providing a data-centric approach to training LLMs in a multi-task manner.

\paragraph{General Large Language Models} 
Given the increased accessibility to large amounts of publicly available data, there has been a constant upward trend in releasing instruct-tuned LLMs. These models include Alpaca \citep{alpaca}, an LLM trained on a dataset augmented with GPT-3. Recently, the latest open-source LLMs, Mistral-v0.3 \citep{jiang2023mistral} and Llama3 \citep{llama3modelcard}, have joined the community, showcasing impressive human-like capabilities across various domains. However, these models are not tailored to any specific domain and often underperform in specialized areas such as finance and ESG.

\paragraph{Financial Large Language Models} 
Financial Large Language Models (FinLLMs) are specifically developed to handle financial text data, offering more precise financial analysis and predictions. One of the earlier efforts, BloombergGPT \citep{wu2023bloomberggpt}, is a 50B model trained on a massive dataset comprising a mixture of financial and general text data. However, it is not publicly accessible and hence there is a call for more open and inclusive alternatives. Other open-source alternatives includes FinGPT \citep{yang2023fingpt, 2023finnlp} and CFGPT \citep{lei2024cfbenchmark}, which introduce tools focused on data acquisition, cleaning, and preprocessing. Their goal is to democratize financial data and the development of FinLLMs, offering a wide range of potential applications. Nonetheless, these efforts have not addressed key concerns on the imbalance in training data and lack of knowledge in the ESG domain. CFGPT faces limitations in language such as only being limited to the Chinese language.
\paragraph{Financial Benchmarks} 
As FinLLMs rapidly advance, the importance of financial evaluation benchmarks has grown significantly. For example, FinGPT Benchmarks and FLUE, focused on assessing NLP tools on a wide array of tasks such as NER and SA. PIXIU \citep{xie2023pixiu, xie2024finben} is a large-scale multitask dataset containing 136K data samples as well as offering benchmarks covering five downstream tasks. However, these evaluation frameworks lack specialized ESG content. We aimed to bridge this gap by introducing TCFD-Bench.

\section{SusGen-GPT}

\subsection{Framework}

\begin{figure*}[ht]
    \centering
    \includegraphics[width=\linewidth]{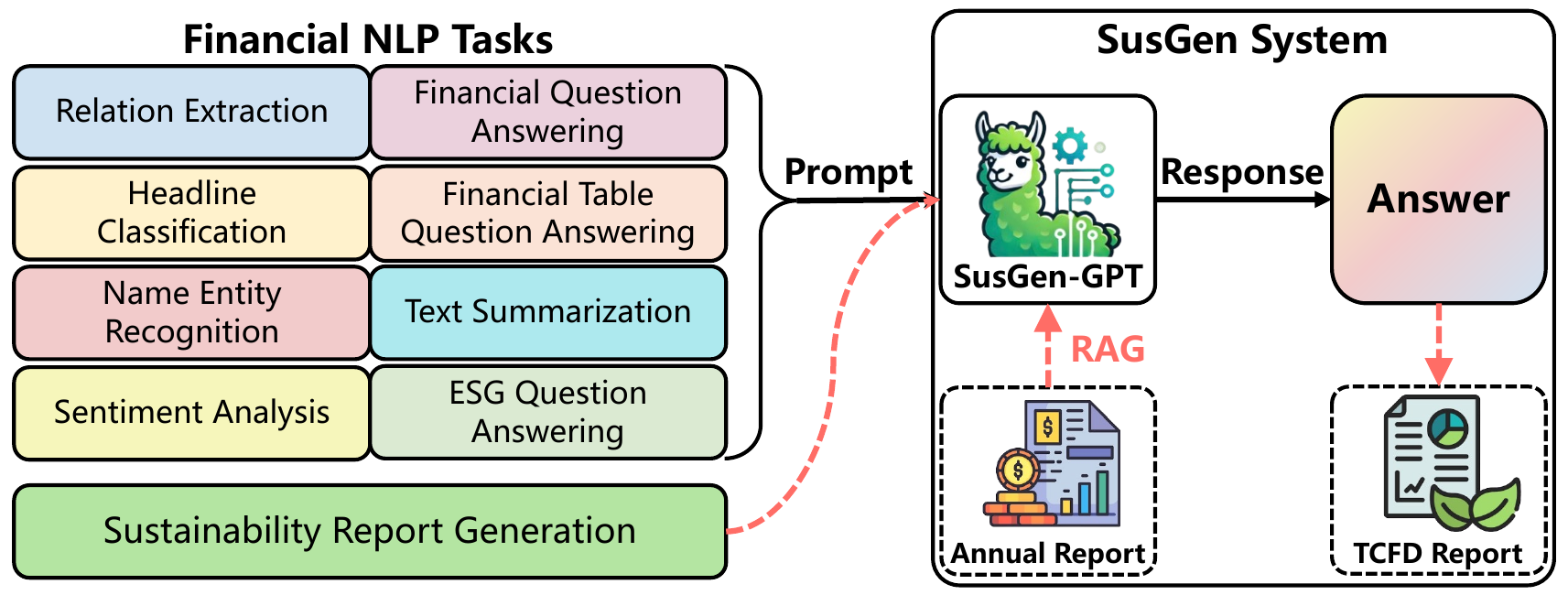}
    \caption{Overview of the SusGen System Pipeline.}
    \label{fig:pipeline}
\end{figure*}

The system, \textit{SusGen}, utilizes SusGen-GPT integrated with Retrieval-Augmented Generation (RAG) specifically for the sustainability report generation task, as shown in Figure \ref{fig:pipeline}. For most financial NLP tasks, SusGen processes the input by prompting and directly feeding it into SusGen-GPT to generate responses. However, for the sustainability report generation task, the system employs RAG to extract relevant information from raw, unstructured annual reports. This extracted information is then combined into pre-defined prompts, which SusGen-GPT uses to generate a comprehensive TCFD-compliant report. The provided summary ensures the generated report adheres to TCFD standards. Additionally, the model is able to answer ESG-related queries concerning company report.

\subsection{Data Construction}
\begin{figure}[ht]
    \centering
    \includegraphics[width=\linewidth]{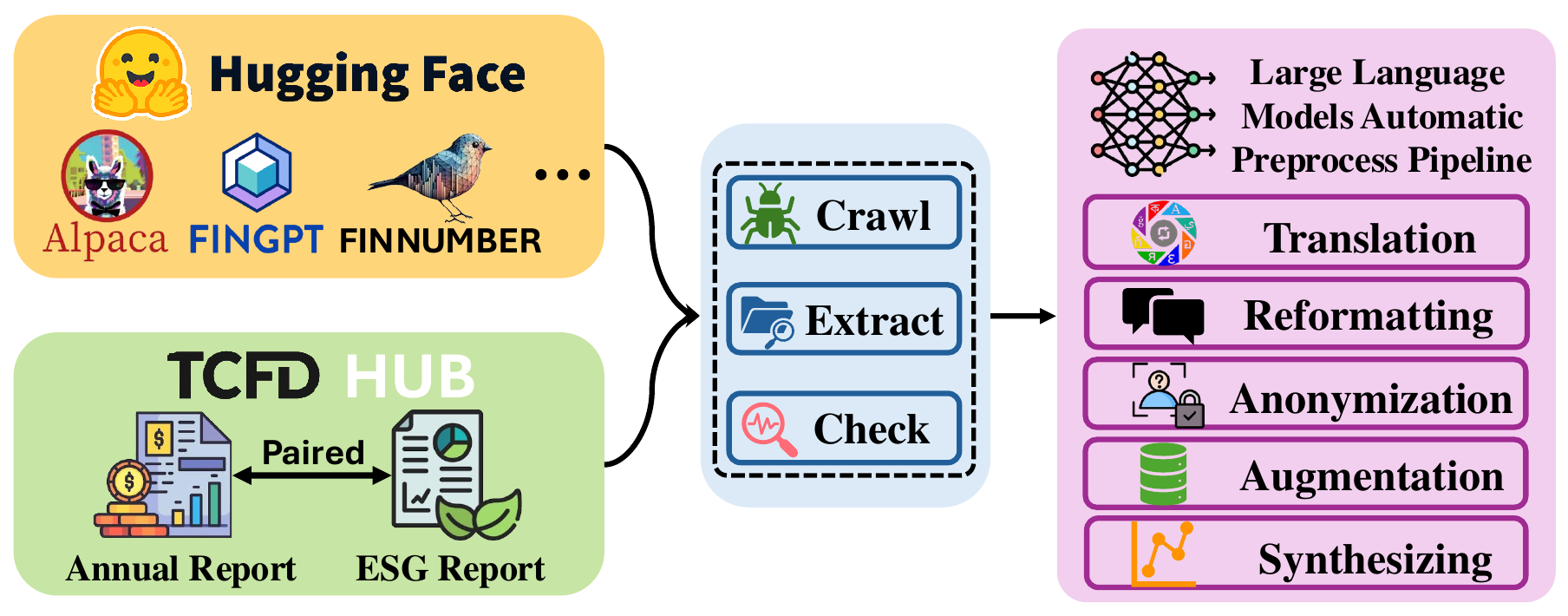}
    \caption{The pipeline of SusGen-30K data construction. The process involves collecting data open-source datasets from hugging-face and company reports from TCFD-Hub Database, followed by quality control and various automatic LLMs pre-processing steps to create the final instruction-following format dataset.} 

    \label{fig:data_construction}
\end{figure}

\noindent\textbf{SusGen-30K}
The dataset is originally sourced from two primary sources: open-source datasets available on Hugging Face\footnote{\url{https://huggingface.co/}} and annual reports sourced from TCFDHub\footnote{\url{https://www.tcfdhub.org/reports}} Database. The construction process for SusGen-30K involves a comprehensive automatic pipeline that starts with data collection from various sources such as company reports (including annual and ESG reports), publicly available financial datasets, and automated content crawlers that scrape financial data from the web, shown as the Figure \ref{fig:data_construction}. 

This raw data undergoes thorough preprocessing steps, including manual annotation to extract useful content, machine-translated data to augment the dataset with multilingual data, and other augmentation techniques to generate novel data samples. We also include anonymization to remove sensitive information and comply with privacy regulations. Finally, the preprocessed data is reformatted into a format compatible with the Supervised-FineTuning(SFT) dataset, ensuring it is well-balanced and ready for training models in financial NLP and Sustainability Report Generation. This structured approach ensures that the dataset is robust, diverse, and high-quality, suitable for advancing the field of sustainable finance. The collected data are then divided into the seven aforementioned financial tasks outlined in Appendix \ref{appendix:2}. Notably, to prevent the model from losing general capabilities, we also incorporated a portion of general and mathematical data into the mix.

Inspired by the scaling law \citep{kaplan2020scaling} and Common-7B \citep{li2024common}, we perform scaling on the dataset to address the imbalance in sample size between the different tasks in the dataset. For large-scale category data, we down-sample them based on data quality to create a well-balanced dataset. Finally, we concatenate all the samples to form the SusGen-30K instruction dataset, which can be used for the financial NLP domain. For detailed information regarding the data sources and composition, please refer to Figure \ref{tab:dataset} in Appendix \ref{appendix:1}.

\begin{figure}[ht]
    \centering
    \includegraphics[width=\linewidth]{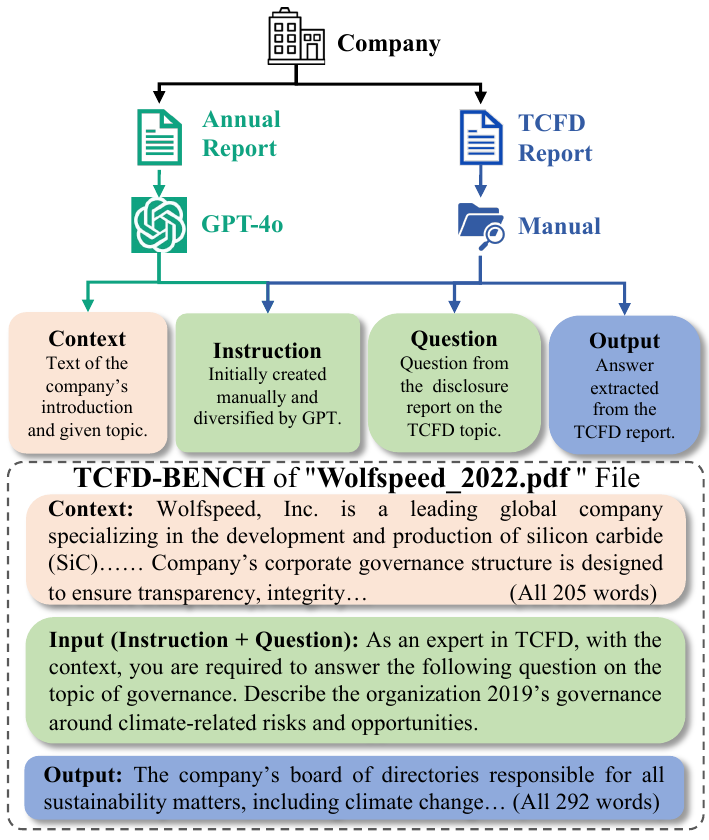}
    \caption{The data construction pipeline of TCFD-Bench, illustrated with an example extracted and processed from the Wolfspeed\_2022.pdf reports.} 

    \label{fig:finben-pipeline}
\end{figure}

\noindent\textbf{TCFD-Bench}
The benchmark includes a balanced distribution of tasks related to ESG reporting, ensuring coverage of key areas like governance, risk management, and strategic planning, which offers significant potential to streamline and enhance the quality of ESG reports in the TCFD format. A sample is shown in Figure \ref{fig:finben-pipeline}, illustrating that each report includes three main parts: context, input, and output. The context section provides a detailed introduction to the company, outlining its specific structure regarding the given topic (governance). The input consists of the instruction and question, while the output presents the answer from the report. All text is extracted using GPT-4o and manual effort, as depicted in Figure \ref{fig:finben-pipeline}. The context information is generated from annual reports using GPT-4o, while the TCFD reports are manually extracted to obtain the questions and answers. To be specific, the explicit question-answer pairs are extracted from 14 ESG PDF reports in the TCFD format. Then we anonymize the data to cover sensitive information, and use Mistral 7B to generate diversified instructions to guide the model's performance. This dual approach leverages both automated large language models and human expertise to build a comprehensive dataset for ESG reporting.

\subsection{Statistics}

\begin{figure}[ht]
    \centering
    \includegraphics[width=\linewidth]{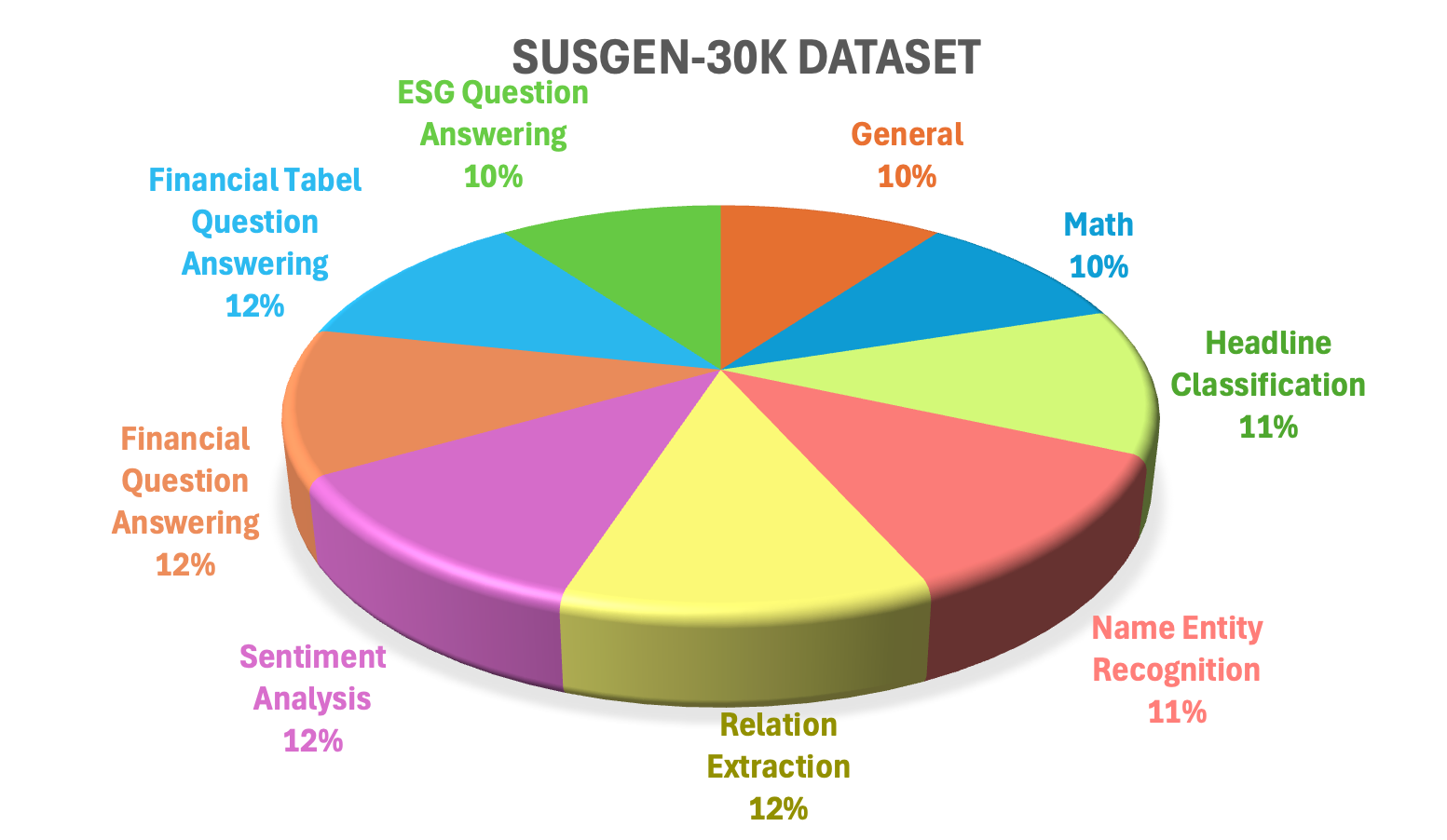}
    \caption{SusGen-30K Category Distribution. Highlight the proportion of data dedicated to each specific task area in financial NLP.}
    \label{fig:susgen-30k-c}
\end{figure}

In this section, we present the statistical information about our training dataset, SusGen-30K. As illustrated in Figure \ref{fig:susgen-30k-c}, the dataset is well-balanced across various task categories, ensuring comprehensive coverage of financial and ESG domains. Specific details regarding the dataset’s categories can be found in the Appendix \ref{appendix:1-data}, including Table \ref{tab:dataset}, which provides information on the data sources and quantities for each category, and Figure \ref{fig:susgen30k_dist}, which shows the token length distribution for each task in the final SusGen-30K dataset.
This balanced distribution allows the model to learn effectively from multi-tasks with low bias, contributing to the robustness and versatility of SusGen-GPT in handling diverse financial and ESG-related tasks.

\subsection{Evaluations}

Our evaluation includes six adapted tasks and two non-adapted (off-the-shelf) tasks, the latter consisting of Text Summarization and Sustainability Report Generation, which were not present in the training dataset. The metrics used to evaluate the performance of SusGen-GPT on various financial and ESG tasks are as follows. 

For Financial Question Answering and Financial Table Question Answering, we employed Exact Match Accuracy and F1 score to measure the precision of the answers. For Headline Classification, we used the Micro F1-score to balance precision and recall across all classes. Named Entity Recognition was assessed using the Entity F1-score to evaluate the accuracy of recognizing and classifying named entities. Sentiment Analysis used the F1-score and Accuracy to measure the balance between precision and recall for predicted sentiments. For Text Summarization, we utilized the ROUGE \citep{lin2004rouge} and BERTScore \citep{zhang2019bertscore} to evaluate the quality of the summaries by comparing the overlap of unigrams between the generated and reference summaries. For sustainable report generation, model performance was evaluated using BERTScore, ROUGE, METEOR \citep{banerjee2005meteor}, and BLEU-N \citep{papineni2002bleu} scores. These metrics were used to measure the similarity of the machine-generated text to the expert reference content, ensuring that the evaluations are robust and reliable.\\ The chosen metrics are relevant and effective in capturing the performance nuances of each task, offering a detailed view of how well the model performs in each area. Evaluation scores were calculated for each sub-task individually to provide specific insights into each task's performance. Additionally, we provide the average score for each category to offer a holistic view of SusGen-GPT's performance across different tasks, highlighting the model’s strengths and areas for improvement.

\section{Experiments \& Analysis}
\subsection{Experimental Setup}
Based on the understanding that LoRA can effectively retain learned knowledge without significant forgetting \citep{biderman2024lora}, we chose the QLoRA \citep{dettmers2024qlora} method over full fine-tuning to preserve the model’s general capabilities while ensuring computational efficiency. Our experiments employed the SusGen-GPT models, leveraging four baseline models: Mistral-v0.3-7B, Mistral-Instruct-v0.3-7B\footnote{The model is released by \href{https://mistral.ai/}{Mistral AI} under the \href{https://choosealicense.com/licenses/apache-2.0/}{Apache 2.0 license} for both commercial and non-commercial usage.}, LLaMA-3-8B, and LLaMA-3-8B-Instruct\footnote{LLaMA3 models are licensed under a bespoke commercial \href{https://llama.meta.com/llama3/license/}{license} by \href{https://llama.meta.com/}{Meta AI}.}, using the QLoRA method for supervised fine-tuning due to its computational efficiency. The experiments were conducted on two NVIDIA RTX 24GB 3090 Ti GPUs. We use different scale datasets of our curated SusGen-30K as the training data. During training, we employ 32-bit Paged AdamW \citep{adamw} optimizer with a cosine learning rate schedule for total 3 epochs of training. The learning rate is set to 2e-5, 10\% warmup steps, 8 batchsize per device with 8 gradient accumulation steps. The maximum token length is set 2048 tokens with alpaca prompt template. And we use 4-bit quantization with double quantization enabled and bfloat16 as the compute data type, set lora rank to 16 and alpha to 32 with a dropout rate of 0.1. Out of twelve models we trained, the most resource-intensive one, involving 30K data records and 8B model, takes around 10 hours on our device.

During evaluation, we use the same alpaca prompt shown in Appendix \ref{appendix:prompt} as used in Training and combining vllm inference optimization techniques. We use LangChain\footnote{https://python.langchain.com/} to manage vector-database retriever. We use all-mpnet-base-v2\footnote{https://huggingface.co/sentence-transformers/all-mpnet-base-v2} for text chunk embedding, split reports into chunks of 1024 tokens and retrieve the top 10 related chunks. We set the temperature to 0.2, top\_p to 0.9, top\_k to 40 and repetition\_penalty to 1.2.

\subsection{Benchmarks \& Baseline Models}
In this section, we introduce the benchmarks which consist of 14 datasets across 8 tasks in financial and esg NLP domain and baseline models used to evaluate SusGen-GPT's performance.

\paragraph{Benchmarks} Text Summarization and Sustainability Report Generation are considered two non-adapted tasks because our training set does not explicitly include them, while the other six tasks are regarded as adapted tasks. For finanical Q\&A task, the FinQA \citep{chen2021finqa} dataset focuses on multi-step numerical reasoning through financial reports. In finanical table Q\&A, the TATQA \citep{zhu2021tat} dataset addresses multi-step numerical reasoning through financial tables, while the ConvFinQA \citep{chen2022convfinqa} dataset involves multiple rounds of Q\&A based on earnings reports and tables. Sentiment analysis examines the linguistic and economic meanings in financial texts, using FinQASA \citep{maia201818} for sentiment extraction and FOMC \citep{shah2023trillion} to categorize sentences as "hawkish" or "dovish". As for news headlines classification, the MultiFin \citep{jorgensen2023multifin} dataset classifies financial texts like analyst reports, news stories, and investor commentary. The MLESG \citep{chen2023multi} dataset detects ESG issues. Named entity recognition extracts entities from financial agreements and SEC documents using NER \citep{alvarado2015domain} and FINER-ORD \citep{shah2023finer} datasets. Relation extraction uses the FINRED \citep{sharma2022finred} dataset to identify relationships in financial news and earnings records, such as "products produced" and "manufacturers." The SC \citep{mariko2020financial} dataset discerns causal relationships in news and SEC filings. For text summarisation, EDTSUM \citep{zhou2021trade} dataset abstracts financial news articles into concise summaries. In sustainability report generation, we utilized the proposed TCFD-Bench, requiring the model to generate TCFD-format ESG reports based on relevant content from company annual reports.

\paragraph{Baseline Models} For closed-source LLMs, we compare our model with OpenAI’s GPT-4 \citep{openai2023gpt4}, which demonstrates exceptional zero-shot performance across multiple NLP tasks, and Gemini \citep{team2023gemini}, a multimodal model capable of processing both text and images, enhancing performance in cross-modal tasks. Among open-source LLMs, we include Mistral 7B-Inst-v0.2/v0.3 \citep{jiang2023mistral}, a high-performing model in the open-source community, and LLaMA3 \citep{llama3modelcard}, Meta’s state-of-the-art model that significantly improves accuracy and efficiency in text generation and comprehension. Additionally, we evaluate FinMA7B \citep{xie2023pixiu}, optimized for financial text analysis, and Falcon7B \citep{almazrouei2023falcon}, both 7B-parameter models designed for specialized and diverse NLP tasks, respectively.

\subsection{Main Results and Comparison}

We evaluate and compare SusGen-GPT on eight tasks in total against other baseline models, including seven financial NLP tasks using well-established benchmarks as well sustainability report generation (SRG) using our proposed TCFD-Bench. The performance of SusGen-GPT across the seven financial tasks is presented in Table \ref{tab:generaltasks} and Figure \ref{fig:result_merge}, while the evaluation results for SRG can be found in Table \ref{tab:tcfdbench}.

\begin{table*}[ht]
		\centering
		\scriptsize
 \renewcommand{\arraystretch}{1.2}
 \scalebox{0.8}{
	\begin{tabular}{llccccccccccccc}
			\toprule
			\textbf{Datasets} 
			&\textbf{Tasks}
                &\textbf{Metrics}
			&\makecell{\textbf{SusGen-}\\\textbf{GPT (Ours)}}
			&\makecell{GPT4\textcolor{black}{*}}
			&\makecell{Gemini\textcolor{black}{*}}
   &\makecell{LLaMA2\\70B}
			&\makecell{LLaMA3\\8B}
			&\makecell{LLaMA3\\8B-Inst}
			&\makecell{FinMA\\7B}
            &\makecell{Falcon\\7B}
            &\makecell{Mistral\\7B-Inst-v0.2}
            &\makecell{Mistral\\7B-Inst-v0.3}

			\\\midrule
   
			\textcolor{black}{FiQASA \citep{maia201818}}
			&SA&F1&0.72&0.70&0.71&0.73&0.72&0.72&0.69&0.67&0.65&\textbf{0.74}\\

            \textcolor{black}{FOMC \citep{shah2023trillion}}
			&SA&F1&\textcolor{black}{0.70}&\textcolor{black}{\textbf{0.71}}&0.53&\textcolor{black}{0.49}&\textcolor{black}{0.53}&\textcolor{black}{0.47}&\textcolor{black}{0.49}&0.30
            &\textcolor{black}{0.30}&\textcolor{black}{0.37}\\\hline
            \textcolor{black}{MultiFin \citep{jorgensen2023multifin}}&HC&MicroF1&0.52&\textbf{0.65}&0.62&0.63&0.50&0.56&0.14&0.09
            &0.50&\textcolor{black}{0.51}\\

            \textcolor{black}{MLESG \citep{chen2023multi}}
			&HC&MicroF1&\textbf{0.51}\textcolor{black}&0.35&0.34&0.31&{0.23}&\textcolor{black}{0.48}&{0.00}&0.06
            &0.47&0.49\\\hline

\textcolor{black}{NER \citep{alvarado2015domain}}
			&NER&EntityF1&0.35&\textbf{0.83}&0.61&0.04&0.06&0.04&0.39&0.00
            &0.17&0.15\\

            \textcolor{black}{FINER-ORD \citep{shah2023finer}}
			&NER&EntityF1&\textcolor{black}{0.18}&\textcolor{black}{\textbf{0.77}}&\textcolor{black}{0.14}&\textcolor{black}{0.07}&\textcolor{black}{0.06}&\textcolor{black}{0.04}&\textcolor{black}{0.00}&0.00
            &\textcolor{black}{0.08}&\textcolor{black}{0.14}\\\hline
    
 \textcolor{black}{FinRED \citep{sharma2022finred}}
			&RE&F1&\textbf{0.23}&0.02&\textcolor{black}{0.00}&\textcolor{black}{0.00}&\textcolor{black}{0.04}&\textcolor{black}{0.08}&\textcolor{black}{0.00}&0.00 
            &0.13&\textcolor{black}{0.14}\\

            \textcolor{black}{SC \citep{mariko2020financial}}
			&RE&F1&\textbf{0.96}&0.81&\textcolor{black}{0.74}&\textcolor{black}{0.61}&\textcolor{black}{0.93}&\textcolor{black}{0.90}&\textcolor{black}{0.19}&0.67
            &0.90&\textcolor{black}{0.85}\\\hline
			FinQA \citep{chen2021finqa}
			&FINQA&EmAcc&0.57&\textbf{0.63}&\textcolor{black}{0.00}&\textcolor{black}{0.06}&0.16&\textcolor{black}{0.54}&\textcolor{black}{0.04}&0.00
            &\textcolor{black}{0.31}&\textcolor{black}{0.32}\\\hline
			 TATQA \citep{zhu2021tat}
			&FINTQA&EmAcc&\textbf{0.80}&0.13&0.18&\textcolor{black}{0.01}&\textcolor{black}{0.26}&\textcolor{black}{0.60}&0.00&0.00
            &0.48&\textcolor{black}{0.52}\\
			ConvFinQA \citep{chen2022convfinqa}
			&FINTQA&EmAcc&0.69&\textbf{0.76}&0.43&\textcolor{black}{0.25}&0.21&\textcolor{black}{0.65}&\textcolor{black}{0.20}&0.00
            &\textcolor{black}{0.48}&\textcolor{black}{0.58}\\\hline
			\multirow{2}{*}{EDTSUM \citep{zhou2021trade}}
			&\multirow{2}{*}{SUM}&Rouge-1&\textcolor{black}{0.27}&\textcolor{black}{0.20}&\textcolor{black}{\textbf{0.39}}&\textcolor{black}{0.25}&\textcolor{black}{0.11}&\textcolor{black}{0.15}&\textcolor{black}{0.13}&0.12
            &\textcolor{black}{0.15}&0.18\\
			& &BertScore&\textcolor{black}{0.54}&\textcolor{black}{0.67}&\textcolor{black}{\textbf{0.72}}&\textcolor{black}{0.68}&\textcolor{black}{0.41}&\textcolor{black}{0.47}&\textcolor{black}{0.38}&0.51
            &\textcolor{black}{0.48}&0.49\\
			\bottomrule
        \end{tabular}}
		\caption{Comparison of zero-shot and few-shot performance between our model and baseline LLMs on seven general financial tasks. ``*'' represents the evaluation result from the previous paper FinBen \citep{xie2024finben}.}	
  	\label{tab:generaltasks}
\end{table*}  

SusGen-GPT demonstrates competitive performance across multiple financial benchmarks. In SA, it achieves an F1 score of 0.72 on the FiQASA dataset, comparable to GPT-4’s 0.70, though GPT-4 slightly outperforms it on the FOMC dataset (0.71 vs. 0.70). For HC, SusGen-GPT scores 0.52 on the MultiFin dataset, trailing GPT-4’s 0.65, but leads on the MLESG dataset with a score of 0.51. In NER, it achieves 0.35 on the NER dataset and 0.18 on the FINER-ORD dataset, but these results fall short of GPT-4’s 0.83 and 0.77, respectively. In RE, SusGen-GPT excels with an F1 of 0.96 on the SC dataset, outperforming all others, though it performs modestly on FinRED (0.23). For FinQA and FinTQA, the model scores 0.57 on FinQA (slightly behind GPT-4’s 0.63) and 0.80 on TATQA, surpassing other models and showcasing strong financial question-answering capabilities.

For sustainability report generation, SusGen-GPT was evaluated on TCFD-Bench against \textsc{ChatReport} (Table \ref{tab:tcfdbench}). SusGen-GPT outperformed \textsc{ChatReport} in Rouge-L (0.20 vs. 0.14), BERTScore (0.40 vs. 0.32), and METEOR (0.27 vs. 0.12), while \textsc{ChatReport} led marginally on BLEU-1 (0.41 vs. 0.39). These highlight SusGen-GPT’s effectiveness in generating ESG reports.

In conclusion, our models achieved near-GPT-4 performance across eight financial tasks, even surpassing it on some, using only 7-8B parameters compared to GPT-4’s 1,700B. This demonstrates the efficiency and effectiveness of our smaller models in achieving state-of-the-art results.

 \begin{table}[ht]
    \centering
    \footnotesize
    \renewcommand{\arraystretch}{1.3}
    \scalebox{0.55}{
    \begin{tabular}{lccccccc}
        \toprule
        \textbf{Models} & \textbf{Rouge-L} & \textbf{Bert-Score} & \textbf{METEOR} & \textbf{BLEU-1} & \textbf{BLEU-2} & \textbf{BLEU-3} & \textbf{BLEU-4} \\
        \midrule
        ChatReport & 0.14 & 0.32 & 0.12 & \textbf{0.41} & 0.10 & 0.03 & 0.02 \\
        \midrule
        \makecell{\textbf{\textsc{SuSGen-}} \\ \textbf{\textsc{GPT} (Ours)}} & \textbf{0.20} & \textbf{0.40} & \textbf{0.27} & 0.32 & \textbf{0.12} & \textbf{0.07} & \textbf{0.05} \\
        \bottomrule
    \end{tabular}}
    \caption{Comparison of sustainability report generation performance on TCFD-Bench between our model and \textsc{ChatReport}.}
    \label{tab:tcfdbench}
\end{table}

\subsection{Ablation Study}

\begin{figure*}[ht]
    \centering
    \includegraphics[width=\linewidth]{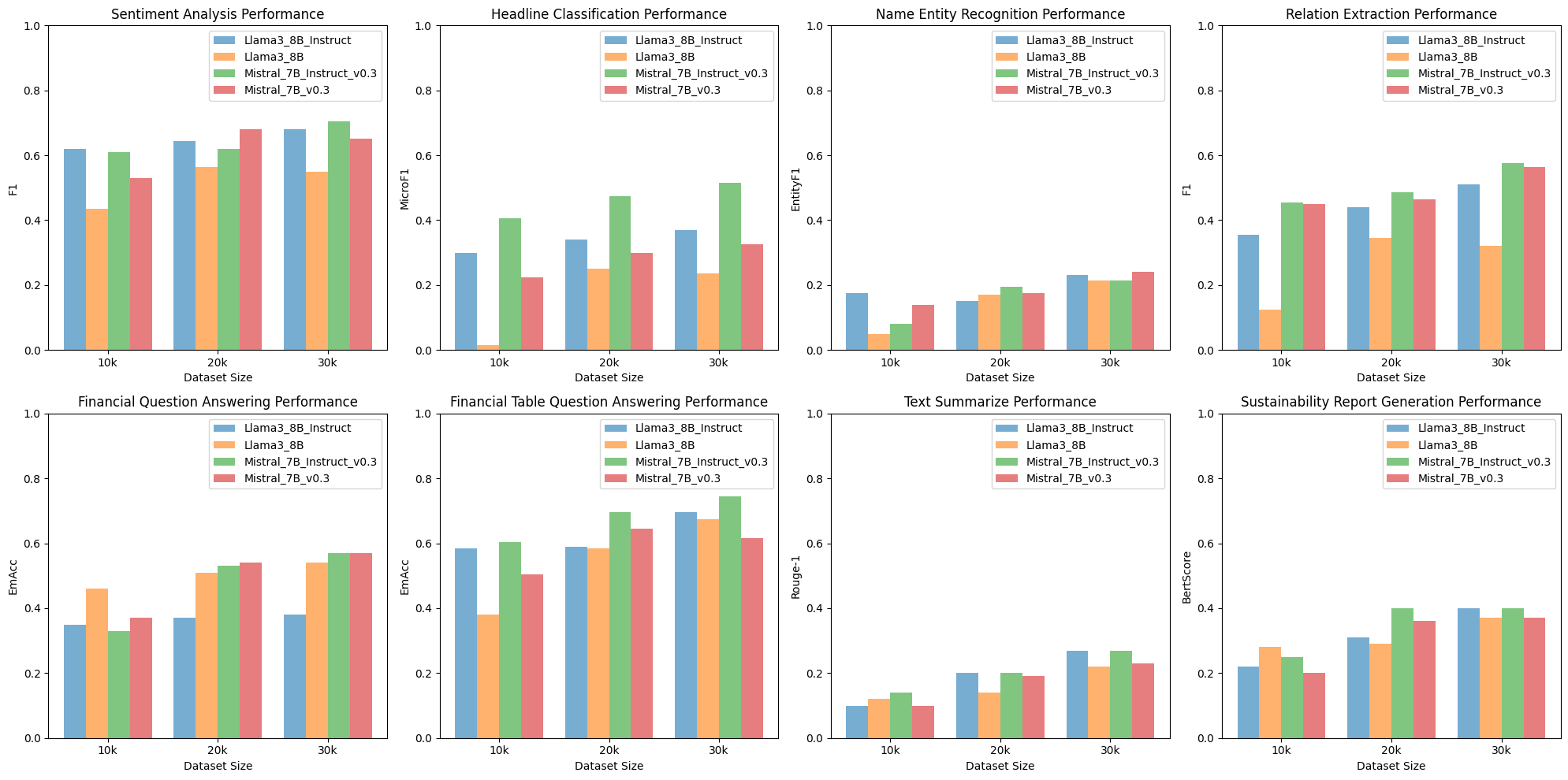}
    \caption{Ablation study results of our models SusGen-GPT trained on 10k, 20k, and 30k subsets of our dataset SusGen-30K, illustrating the data scaling effect across eight financial NLP tasks.}
    \label{fig:dataset_scale}
\end{figure*}

In this section, we investigate the effect of dataset scaling on SusGen-GPT across eight financial NLP tasks using datasets of 10k, 20k, and 30k samples. Comprehensive results can be found in Table \ref{tab:datascale} in Appendix \ref{appendix:3}, and the performance trends for all tasks are illustrated in Figure \ref{fig:dataset_scale}.

Overall, the results show a clear trend: increasing the dataset size consistently improves performance across all tasks. This pattern is evident as larger datasets allow the models to better capture complex patterns in financial and ESG-related data, leading to higher scores in various metrics such as F1 and MicroF1.

For instantce, in FiQASA, SusGen-GPT reaches an F1 score of 0.72 at the 30k scale, up from 0.66 at 10k. Similarly, in the FOMC dataset, SusGen-GPT achieves an F1 score of 0.70 at 30k, compared to 0.61 at 10k. In Relation Extraction tasks like FinRED, the model improves from an F1 score of 0.04 at 10k to 0.23 at 30k, showing substantial performance gains as the dataset grows. Likewise, in NER tasks, the EntityF1 score increases from 0.27 at 10k to 0.35 at 30k. These findings clearly demonstrate that increasing the dataset size enables SusGen-GPT to capture more complex patterns, leading to better overall performance in financial NLP tasks.

\begin{figure*}[ht]
\centering
\includegraphics[height=0.33\textheight]{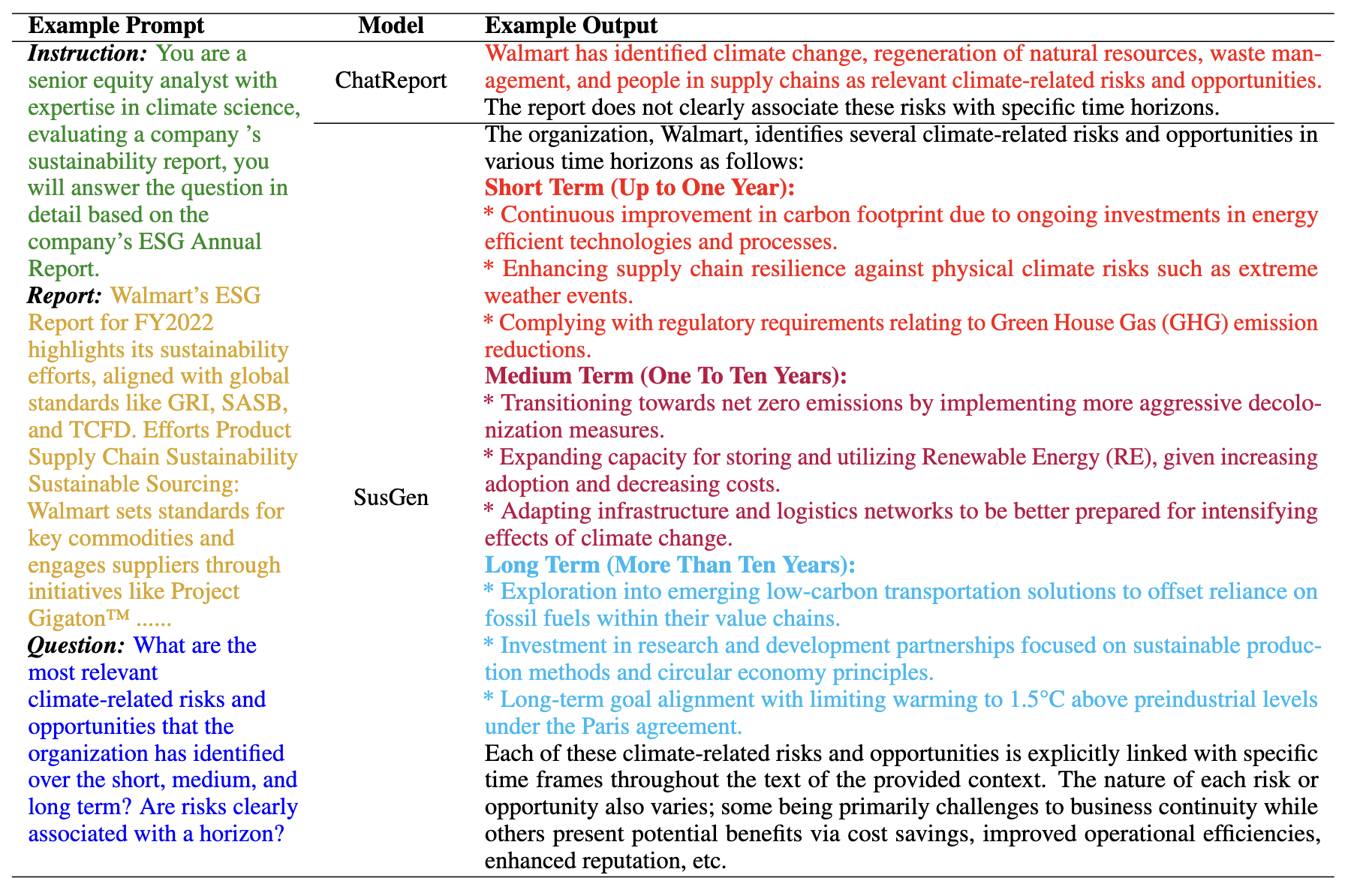}

\caption{\label{tab:prompt_dev} A generated Sustainability Report Example. A comparison of the output obtained by our SusGen-GPT and ChatReport with the same question in the TCFD-Bench.}
\end{figure*}

\subsection{Qualitative Analysis}

\noindent\textbf{Generated Sustainability Report Analysis} A comparison of the outputs obtained by SusGen-GPT and ChatReport based on the company's annual report after inputting the same TCFD standard ESG-related questions is shown in Appendix \ref{appendix:4}. The detailed comparison of these models is illustrated in Figure \ref{tab:prompt_dev}. ChatReport is a robust, publicly available text generation model for creating ESG reports in the TCFD format, leveraging the capabilities of GPT-3.5. However, SusGen-GPT surpasses ChatReport in several key areas. It offers a detailed breakdown, specific action points, enhanced contextual clarity, and a comprehensive approach. 

SusGen-GPT's structured format provides significant advantages, making it a superior tool for generating TCFD-format ESG reports. These improvements ensure that reports produced by SusGen-GPT are not only more informative but also more actionable and easier to understand. By breaking down the risks and opportunities into specific time horizons, SusGen-GPT provides a clear roadmap for addressing climate-related challenges and leveraging opportunities. Each identified risk and opportunity is linked to specific actions and goals, ensuring that the report is practical and aligned with the company’s strategic objectives. The enhanced contextual clarity provided by SusGen-GPT allows users to better understand the implications of each risk and opportunity, aiding stakeholders in making informed decisions. The detailed action points help in formulating concrete sustainability strategies, improving operational efficiency, enhancing reputation, and ensuring regulatory compliance.

In contrast, while ChatReport provides a broad overview of climate-related risks and opportunities, it lacks the depth and specificity found in SusGen-GPT’s output. ChatReport's responses are less structured and do not consistently associate risks with specific time horizons, which can make it harder for users to prioritize actions and understand the timeline for implementation. In summary, SusGen-GPT's comprehensive and structured approach to generating ESG reports in the TCFD format makes it a more effective tool for companies aiming to address climate-related risks and opportunities in a clear, actionable, and strategically aligned manner.

\section{Conclusion and Future Work}
In conclusion, our work specifically addresses the significant gap in specialized language models for the financial and ESG sectors by introducing SusGen-30K and SusGen-GPT. The balanced nature of SusGen-30K allows SusGen-GPT to excel in multiple downstream financial tasks including sustainability report generation, achieving superior performance even than other larger scale LLMs. Additionally, our proposed TCFD-Bench benchmark provides a rigorous evaluation framework for ESG report generation. Future work will focus on expanding the dataset to cover more specialized tasks within the ESG domain, enhancing the model's capabilities, and refining the benchmark to include more diverse evaluation metrics. This work lays a robust foundation for developing advanced tools that meet the growing demands of the financial sector and contribute to climate-related financial disclosures.

\clearpage

\section*{Limitations}
\paragraph{Limited Model Performance}
One of the limitations of our work is the performance of our models, which is influenced by resource constraints. The large models we utilize are typically in the range of 7B/8B in terms of parameters. Due to limited resources, we have not had the opportunity to explore the potential benefits of 70B parameter or larger models. As a result, our models may not achieve their full potential performance, and this is an aspect we are mindful of as we continue our work. We aim to address these limitations in the future and strive to improve the performance of our models as resources allow.

\paragraph{Limited Evaluation from Expert}
While experts have conducted quality analysis for certain cases, the evaluation of large-scale data relies on automated scores such as BLEU and ROUGE metrics. These automated evaluations, while valuable, may potentially introduce biases and lack the nuanced understanding that human expertise provides, particularly in the ESG (Environmental, Social, and Governance) domain. Furthermore, there is a shortage of expert resources in the ESG domain, which limits the comprehensive evaluation of the data concerning ESG factors. As a result, the evaluation may not fully capture the depth and complexity of the ESG-related aspects of the data.

\paragraph{Unsuitable for various ESG Subtasks}
The model's performance may exhibit significant variability across different subfields, highlighting the necessity for targeted research and optimization for specific ESG-related subtasks. For example, certain subfields, such as renewable energy investment or sustainable supply chain management, might require more customized approaches to ensure the model's performance meets the expected standards. These variations necessitate a more granular understanding of each subfield's unique characteristics and requirements, demanding further data gathering and model adjustments.

\paragraph{Insufficient Diverse Report Template}
Despite the significant progress achieved by our model, there are still notable limitations concerning the comprehensiveness of the dataset. Firstly, our model was predominantly trained on a limited number of reports in the TCFD (Task Force on Climate-related Financial Disclosures) format, with sparse representation from other key standards such as the GRI (Global Reporting Initiative), SASB (Sustainability Accounting Standards Board), and CDP (Carbon Disclosure Project). This narrow data source limits the model's generalization capabilities and may hinder its performance when dealing with reports adhering to different standards and formats.

\section*{Ethical Considerations}
\paragraph{False Information from LLMs}
One of the pressing issues in this work is the phenomenon of model hallucination, where the model generates information that is not present in the input data. This problem is particularly significant when applying advanced techniques like LLMs to financial data, as generating false information can have serious implications. For instance, inaccurate or misleading financial reports can lead to incorrect business decisions, regulatory non-compliance, and loss of stakeholder trust. Addressing this issue requires ongoing research and development to improve the reliability and accuracy of NLP models.

\paragraph{Bias towards Firm Perspective}
Another consideration is the inherent bias towards the firm's perspective in the extracted information from corporate sustainability reports. This bias arises because the data predominantly originates from the companies themselves, potentially leading to a one-sided view that may overlook critical aspects such as stakeholder opinions and third-party assessments. To mitigate this, future work should aim to incorporate a more diverse set of data sources, including independent audits and reports from non-governmental organizations (NGOs), to provide a more balanced view of corporate sustainability practices.

\paragraph{License of the Tool}
To ensure that our tool is accessible and adaptable by all stakeholders, we have chosen to release it under the Apache License 2.0 later. This open-source license allows for wide distribution, usage, and modification of the tool, thereby facilitating collaborative development and continuous improvement. By doing so, we aim to foster a community of practice that can collectively address the challenges and leverage the opportunities presented by the use of NLP in financial contexts. This approach not only democratizes access to advanced technologies but also encourages transparency and accountability in their application.

\clearpage
\bibliography{ref/anthology,ref/wql,ref/xxn,ref/hhj}

\begin{thebibliography}{50}
\providecommand{\natexlab}[1]{#1}

\bibitem[{Aguda et~al.(2024)Aguda, Siddagangappa, Kochkina, Kaur, Wang, and Smiley}]{aguda-etal-2024-large}
Toyin~D. Aguda, Suchetha Siddagangappa, Elena Kochkina, Simerjot Kaur, Dongsheng Wang, and Charese Smiley. 2024.
\newblock \href {https://aclanthology.org/2024.lrec-main.885} {Large language models as financial data annotators: A study on effectiveness and efficiency}.
\newblock In \emph{Proceedings of the 2024 Joint International Conference on Computational Linguistics, Language Resources and Evaluation (LREC-COLING 2024)}, pages 10124--10145, Torino, Italia. ELRA and ICCL.

\bibitem[{Almazrouei et~al.(2023)Almazrouei, Alobeidli, Alshamsi, Cappelli, Cojocaru, Debbah, Goffinet, Hesslow, Launay, Malartic et~al.}]{almazrouei2023falcon}
Ebtesam Almazrouei, Hamza Alobeidli, Abdulaziz Alshamsi, Alessandro Cappelli, Ruxandra Cojocaru, M{\'e}rouane Debbah, {\'E}tienne Goffinet, Daniel Hesslow, Julien Launay, Quentin Malartic, et~al. 2023.
\newblock The falcon series of open language models.
\newblock \emph{arXiv preprint arXiv:2311.16867}.

\bibitem[{Alvarado et~al.(2015)Alvarado, Verspoor, and Baldwin}]{alvarado2015domain}
Julio Cesar~Salinas Alvarado, Karin Verspoor, and Timothy Baldwin. 2015.
\newblock Domain adaption of named entity recognition to support credit risk assessment.
\newblock In \emph{Proceedings of the Australasian Language Technology Association Workshop 2015}, pages 84--90.

\bibitem[{Banerjee and Lavie(2005)}]{banerjee2005meteor}
Satanjeev Banerjee and Alon Lavie. 2005.
\newblock Meteor: An automatic metric for mt evaluation with improved correlation with human judgments.
\newblock In \emph{Proceedings of the acl workshop on intrinsic and extrinsic evaluation measures for machine translation and/or summarization}, pages 65--72.

\bibitem[{Biderman et~al.(2024)Biderman, Ortiz, Portes, Paul, Greengard, Jennings, King, Havens, Chiley, Frankle et~al.}]{biderman2024lora}
Dan Biderman, Jose~Gonzalez Ortiz, Jacob Portes, Mansheej Paul, Philip Greengard, Connor Jennings, Daniel King, Sam Havens, Vitaliy Chiley, Jonathan Frankle, et~al. 2024.
\newblock Lora learns less and forgets less.
\newblock \emph{arXiv preprint arXiv:2405.09673}.

\bibitem[{Bronzini et~al.(2024)Bronzini, Nicolini, Lepri, Passerini, and Staiano}]{bronzini2024glitter}
Marco Bronzini, Carlo Nicolini, Bruno Lepri, Andrea Passerini, and Jacopo Staiano. 2024.
\newblock Glitter or gold? deriving structured insights from sustainability reports via large language models.
\newblock \emph{EPJ Data Science}, 13(1):41.

\bibitem[{Brown et~al.(2020)Brown, Mann, Ryder, Subbiah, Kaplan, Dhariwal, Neelakantan, Shyam, Sastry, Askell et~al.}]{brown2020language}
Tom Brown, Benjamin Mann, Nick Ryder, Melanie Subbiah, Jared~D Kaplan, Prafulla Dhariwal, Arvind Neelakantan, Pranav Shyam, Girish Sastry, Amanda Askell, et~al. 2020.
\newblock Language models are few-shot learners.
\newblock \emph{Advances in neural information processing systems}, 33:1877--1901.

\bibitem[{Chen et~al.(2023)Chen, Tseng, Kang, Lhuissier, Day, Tu, and Chen}]{chen2023multi}
Chung-Chi Chen, Yu-Min Tseng, Juyeon Kang, Ana{\"\i}s Lhuissier, Min-Yuh Day, Teng-Tsai Tu, and Hsin-Hsi Chen. 2023.
\newblock Multi-lingual esg issue identification.
\newblock In \emph{Proceedings of the Fifth Workshop on Financial Technology and Natural Language Processing and the Second Multimodal AI For Financial Forecasting}, pages 111--115.

\bibitem[{Chen et~al.(2021)Chen, Chen, Smiley, Shah, Borova, Langdon, Moussa, Beane, Huang, Routledge et~al.}]{chen2021finqa}
Zhiyu Chen, Wenhu Chen, Charese Smiley, Sameena Shah, Iana Borova, Dylan Langdon, Reema Moussa, Matt Beane, Ting-Hao Huang, Bryan~R Routledge, et~al. 2021.
\newblock Finqa: A dataset of numerical reasoning over financial data.
\newblock In \emph{Proceedings of the 2021 Conference on Empirical Methods in Natural Language Processing}, pages 3697--3711.

\bibitem[{Chen et~al.(2022)Chen, Li, Smiley, Ma, Shah, and Wang}]{chen2022convfinqa}
Zhiyu Chen, Shiyang Li, Charese Smiley, Zhiqiang Ma, Sameena Shah, and William~Yang Wang. 2022.
\newblock \href {https://arxiv.org/abs/2210.03849} {Convfinqa: Exploring the chain of numerical reasoning in conversational finance question answering}.
\newblock \emph{Preprint}, arXiv:2210.03849.

\bibitem[{Cobbe et~al.(2021)Cobbe, Kosaraju, Bavarian, Chen, Jun, Kaiser, Plappert, Tworek, Hilton, Nakano, Hesse, and Schulman}]{cobbe2021gsm8k}
Karl Cobbe, Vineet Kosaraju, Mohammad Bavarian, Mark Chen, Heewoo Jun, Lukasz Kaiser, Matthias Plappert, Jerry Tworek, Jacob Hilton, Reiichiro Nakano, Christopher Hesse, and John Schulman. 2021.
\newblock Training verifiers to solve math word problems.
\newblock \emph{arXiv preprint arXiv:2110.14168}.

\bibitem[{Dettmers et~al.(2024)Dettmers, Pagnoni, Holtzman, and Zettlemoyer}]{dettmers2024qlora}
Tim Dettmers, Artidoro Pagnoni, Ari Holtzman, and Luke Zettlemoyer. 2024.
\newblock Qlora: Efficient finetuning of quantized llms.
\newblock \emph{Advances in Neural Information Processing Systems}, 36.

\bibitem[{Dubey and et. al(2024)}]{llama3modelcard}
Abhimanyu Dubey and Abhinav~Jauhri et. al. 2024.
\newblock \href {https://arxiv.org/abs/2407.21783} {The llama 3 herd of models}.
\newblock \emph{Preprint}, arXiv:2407.21783.

\bibitem[{Jiang et~al.(2023)Jiang, Sablayrolles, Mensch, Bamford, Chaplot, de~las Casas, Bressand, Lengyel, Lample, Saulnier, Lavaud, Lachaux, Stock, Scao, Lavril, Wang, Lacroix, and Sayed}]{jiang2023mistral}
Albert~Q. Jiang, Alexandre Sablayrolles, Arthur Mensch, Chris Bamford, Devendra~Singh Chaplot, Diego de~las Casas, Florian Bressand, Gianna Lengyel, Guillaume Lample, Lucile Saulnier, Lélio~Renard Lavaud, Marie-Anne Lachaux, Pierre Stock, Teven~Le Scao, Thibaut Lavril, Thomas Wang, Timothée Lacroix, and William~El Sayed. 2023.
\newblock \href {https://arxiv.org/abs/2310.06825} {Mistral 7b}.

\bibitem[{J{\o}rgensen et~al.(2023)J{\o}rgensen, Brandt, Hartmann, Dai, Igel, and Elliott}]{jorgensen2023multifin}
Rasmus J{\o}rgensen, Oliver Brandt, Mareike Hartmann, Xiang Dai, Christian Igel, and Desmond Elliott. 2023.
\newblock Multifin: A dataset for multilingual financial nlp.
\newblock In \emph{Findings of the Association for Computational Linguistics: EACL 2023}, pages 864--879.

\bibitem[{Kaplan et~al.(2020)Kaplan, McCandlish, Henighan, Brown, Chess, Child, Gray, Radford, Wu, and Amodei}]{kaplan2020scaling}
Jared Kaplan, Sam McCandlish, Tom Henighan, Tom~B. Brown, Benjamin Chess, Rewon Child, Scott Gray, Alec Radford, Jeffrey Wu, and Dario Amodei. 2020.
\newblock \href {https://arxiv.org/abs/2001.08361} {Scaling laws for neural language models}.
\newblock \emph{Preprint}, arXiv:2001.08361.

\bibitem[{Lei et~al.(2024)Lei, Li, Cheng, Ding, and Jiang}]{lei2024cfbenchmark}
Yang Lei, Jiangtong Li, Dawei Cheng, Zhijun Ding, and Changjun Jiang. 2024.
\newblock \href {https://arxiv.org/abs/2311.05812} {Cfbenchmark: Chinese financial assistant benchmark for large language model}.
\newblock \emph{Preprint}, arXiv:2311.05812.

\bibitem[{Li et~al.(2024)Li, Wang, Hu, Wei, Zheng, Hu, Zhang, and Peng}]{li2024common}
Chen Li, Weiqi Wang, Jingcheng Hu, Yixuan Wei, Nanning Zheng, Han Hu, Zheng Zhang, and Houwen Peng. 2024.
\newblock \href {https://arxiv.org/abs/2403.04706} {Common 7b language models already possess strong math capabilities}.
\newblock \emph{Preprint}, arXiv:2403.04706.

\bibitem[{Lin(2004)}]{lin2004rouge}
Chin-Yew Lin. 2004.
\newblock Rouge: A package for automatic evaluation of summaries.
\newblock In \emph{Text summarization branches out}, pages 74--81.

\bibitem[{Liu et~al.(2023)Liu, Wang, Yang, and Zha}]{2023finnlp}
Xiao-Yang Liu, Guoxuan Wang, Hongyang Yang, and Daochen Zha. 2023.
\newblock Data-centric fingpt: Democratizing internet-scale data for financial large language models.
\newblock \emph{NeurIPS Workshop on Instruction Tuning and Instruction Following}.

\bibitem[{Loshchilov and Hutter(2019)}]{adamw}
Ilya Loshchilov and Frank Hutter. 2019.
\newblock \href {https://arxiv.org/abs/1711.05101} {Decoupled weight decay regularization}.
\newblock \emph{Preprint}, arXiv:1711.05101.

\bibitem[{Luccioni et~al.(2020)Luccioni, Baylor, and Duchene}]{luccioni2020analyzing}
Alexandra Luccioni, Emily Baylor, and Nicolas Duchene. 2020.
\newblock \href {https://arxiv.org/abs/2011.08073} {Analyzing sustainability reports using natural language processing}.
\newblock \emph{Preprint}, arXiv:2011.08073.

\bibitem[{Maia et~al.(2018)Maia, Handschuh, Freitas, Davis, McDermott, Zarrouk, and Balahur}]{maia201818}
Macedo Maia, Siegfried Handschuh, Andr{\'e} Freitas, Brian Davis, Ross McDermott, Manel Zarrouk, and Alexandra Balahur. 2018.
\newblock Www'18 open challenge: financial opinion mining and question answering.
\newblock In \emph{Companion proceedings of the the web conference 2018}, pages 1941--1942.

\bibitem[{Mariko et~al.(2020)Mariko, Akl, Labidurie, Durfort, De~Mazancourt, and El-Haj}]{mariko2020financial}
Dominique Mariko, Hanna~Abi Akl, Estelle Labidurie, Stephane Durfort, Hugues De~Mazancourt, and Mahmoud El-Haj. 2020.
\newblock Financial document causality detection shared task (fincausal 2020).
\newblock \emph{arXiv preprint arXiv:2012.02505}.

\bibitem[{Masson and Paroubek(2024)}]{masson-paroubek-2024-evaluating}
Corentin Masson and Patrick Paroubek. 2024.
\newblock \href {https://aclanthology.org/2024.lrec-main.578} {Evaluating topic model on asymmetric and multi-domain financial corpus}.
\newblock In \emph{Proceedings of the 2024 Joint International Conference on Computational Linguistics, Language Resources and Evaluation (LREC-COLING 2024)}, pages 6515--6529, Torino, Italia. ELRA and ICCL.

\bibitem[{Ni et~al.(2023)Ni, Bingler, Colesanti-Senni, Kraus, Gostlow, Schimanski, Stammbach, Ashraf~Vaghefi, Wang, Webersinke, Wekhof, Yu, and Leippold}]{ni-etal-2023-chatreport}
Jingwei Ni, Julia Bingler, Chiara Colesanti-Senni, Mathias Kraus, Glen Gostlow, Tobias Schimanski, Dominik Stammbach, Saeid Ashraf~Vaghefi, Qian Wang, Nicolas Webersinke, Tobias Wekhof, Tingyu Yu, and Markus Leippold. 2023.
\newblock \href {https://doi.org/10.18653/v1/2023.emnlp-demo.3} {{CHATREPORT}: Democratizing sustainability disclosure analysis through {LLM}-based tools}.
\newblock In \emph{Proceedings of the 2023 Conference on Empirical Methods in Natural Language Processing: System Demonstrations}, pages 21--51, Singapore. Association for Computational Linguistics.

\bibitem[{OpenAI(2023{\natexlab{a}})}]{gpt4techreport}
OpenAI. 2023{\natexlab{a}}.
\newblock \href {https://cdn.openai.com/papers/gpt-4.pdf} {Gpt-4 technical report}.
\newblock Technical report, OpenAI.

\bibitem[{OpenAI(2023{\natexlab{b}})}]{openai2023gpt4}
OpenAI. 2023{\natexlab{b}}.
\newblock \href {https://arxiv.org/abs/2303.08774} {Gpt-4 technical report}.
\newblock \emph{Preprint}, arXiv:2303.08774.

\bibitem[{OpenAI(2023{\natexlab{c}})}]{instructgpt}
OpenAI. 2023{\natexlab{c}}.
\newblock Instructgpt: Ai for generating instructions.
\newblock https://openai.com/research/instructgpt/.

\bibitem[{Ouyang et~al.(2022)Ouyang, Wu, Jiang, Almeida, Wainwright, Mishkin, Zhang, Agarwal, Slama, Ray et~al.}]{ouyang2022training}
Long Ouyang, Jeffrey Wu, Xu~Jiang, Diogo Almeida, Carroll Wainwright, Pamela Mishkin, Chong Zhang, Sandhini Agarwal, Katarina Slama, Alex Ray, et~al. 2022.
\newblock Training language models to follow instructions with human feedback.
\newblock \emph{Advances in Neural Information Processing Systems}, 35:27730--27744.

\bibitem[{Papineni et~al.(2002)Papineni, Roukos, Ward, and Zhu}]{papineni2002bleu}
Kishore Papineni, Salim Roukos, Todd Ward, and Wei-Jing Zhu. 2002.
\newblock Bleu: a method for automatic evaluation of machine translation.
\newblock In \emph{Proceedings of the 40th annual meeting of the Association for Computational Linguistics}, pages 311--318.

\bibitem[{Shah et~al.(2023{\natexlab{a}})Shah, Paturi, and Chava}]{shah2023trillion}
Agam Shah, Suvan Paturi, and Sudheer Chava. 2023{\natexlab{a}}.
\newblock \href {https://doi.org/10.18653/v1/2023.acl-long.368} {Trillion dollar words: A new financial dataset, task {\&} market analysis}.
\newblock In \emph{Proceedings of the 61st Annual Meeting of the Association for Computational Linguistics (Volume 1: Long Papers)}, pages 6664--6679, Toronto, Canada. Association for Computational Linguistics.

\bibitem[{Shah et~al.(2023{\natexlab{b}})Shah, Vithani, Gullapalli, and Chava}]{shah2023finer}
Agam Shah, Ruchit Vithani, Abhinav Gullapalli, and Sudheer Chava. 2023{\natexlab{b}}.
\newblock Finer: Financial named entity recognition dataset and weak-supervision model.
\newblock \emph{arXiv preprint arXiv:2302.11157}.

\bibitem[{Shah et~al.(2022)Shah, Chawla, Eidnani, Shah, Du, Chava, Raman, Smiley, Chen, and Yang}]{shah2022flue}
Raj~Sanjay Shah, Kunal Chawla, Dheeraj Eidnani, Agam Shah, Wendi Du, Sudheer Chava, Natraj Raman, Charese Smiley, Jiaao Chen, and Diyi Yang. 2022.
\newblock \href {https://arxiv.org/abs/2211.00083} {When flue meets flang: Benchmarks and large pre-trained language model for financial domain}.
\newblock \emph{Preprint}, arXiv:2211.00083.

\bibitem[{Sharma et~al.(2022)Sharma, Nayak, Bose, Meena, Dasgupta, Ganguly, and Goyal}]{sharma2022finred}
Soumya Sharma, Tapas Nayak, Arusarka Bose, Ajay~Kumar Meena, Koustuv Dasgupta, Niloy Ganguly, and Pawan Goyal. 2022.
\newblock Finred: A dataset for relation extraction in financial domain.
\newblock In \emph{Companion Proceedings of the Web Conference 2022}, pages 595--597.

\bibitem[{Taori et~al.(2023)Taori, Gulrajani, Zhang, Dubois, Li, Guestrin, Liang, and Hashimoto}]{alpaca}
Rohan Taori, Ishaan Gulrajani, Tianyi Zhang, Yann Dubois, Xuechen Li, Carlos Guestrin, Percy Liang, and Tatsunori~B. Hashimoto. 2023.
\newblock Stanford alpaca: An instruction-following llama model.
\newblock \url{https://github.com/tatsu-lab/stanford_alpaca}.

\bibitem[{Team et~al.(2023)Team, Anil, Borgeaud, Wu, Alayrac, Yu, Soricut, Schalkwyk, Dai, Hauth et~al.}]{team2023gemini}
Gemini Team, Rohan Anil, Sebastian Borgeaud, Yonghui Wu, Jean-Baptiste Alayrac, Jiahui Yu, Radu Soricut, Johan Schalkwyk, Andrew~M Dai, Anja Hauth, et~al. 2023.
\newblock Gemini: a family of highly capable multimodal models.
\newblock \emph{arXiv preprint arXiv:2312.11805}.

\bibitem[{Touvron and et. al.(2023{\natexlab{a}})}]{touvron2023llama}
Hugo Touvron and Lavril et. al. 2023{\natexlab{a}}.
\newblock Llama: Open and efficient foundation language models.
\newblock \emph{arXiv preprint arXiv:2302.13971}.

\bibitem[{Touvron and et. al.(2023{\natexlab{b}})}]{touvron2023llama2}
Hugo Touvron and Louis~Martin et. al. 2023{\natexlab{b}}.
\newblock \href {https://arxiv.org/abs/2307.09288} {Llama 2: Open foundation and fine-tuned chat models}.
\newblock \emph{Preprint}, arXiv:2307.09288.

\bibitem[{Wang et~al.(2023{\natexlab{a}})Wang, Yang, and Wang}]{wang2023fingptbenchmark}
Neng Wang, Hongyang Yang, and Christina~Dan Wang. 2023{\natexlab{a}}.
\newblock Fingpt: Instruction tuning benchmark for open-source large language models in financial datasets.
\newblock \emph{NeurIPS Workshop on Instruction Tuning and Instruction Following}.

\bibitem[{Wang et~al.(2023{\natexlab{b}})Wang, Yang, and Wang}]{wang2023fingpt}
Neng Wang, Hongyang Yang, and Christina~Dan Wang. 2023{\natexlab{b}}.
\newblock \href {https://arxiv.org/abs/2310.04793} {Fingpt: Instruction tuning benchmark for open-source large language models in financial datasets}.
\newblock \emph{Preprint}, arXiv:2310.04793.

\bibitem[{Wu et~al.(2023)Wu, Irsoy, Lu, Dabravolski, Dredze, Gehrmann, Kambadur, Rosenberg, and Mann}]{wu2023bloomberggpt}
Shijie Wu, Ozan Irsoy, Steven Lu, Vadim Dabravolski, Mark Dredze, Sebastian Gehrmann, Prabhanjan Kambadur, David Rosenberg, and Gideon Mann. 2023.
\newblock \href {https://arxiv.org/abs/2303.17564} {Bloomberggpt: A large language model for finance}.
\newblock \emph{Preprint}, arXiv:2303.17564.

\bibitem[{Xie and et. al.(2023)}]{xie2023pixiu}
Qianqian Xie and Weiguang~Han et. al. 2023.
\newblock \href {https://arxiv.org/abs/2306.05443} {Pixiu: A large language model, instruction data and evaluation benchmark for finance}.
\newblock \emph{Preprint}, arXiv:2306.05443.

\bibitem[{Xie and et. al.(2024)}]{xie2024finben}
Qianqian Xie and Weiguang~Han et. al. 2024.
\newblock \href {https://arxiv.org/abs/2402.12659} {The finben: An holistic financial benchmark for large language models}.
\newblock \emph{Preprint}, arXiv:2402.12659.

\bibitem[{Yang et~al.(2023)Yang, Liu, and Wang}]{yang2023fingpt}
Hongyang Yang, Xiao-Yang Liu, and Christina~Dan Wang. 2023.
\newblock Fingpt: Open-source financial large language models.
\newblock \emph{FinLLM Symposium at IJCAI 2023}.

\bibitem[{Yeo et~al.(2024)Yeo, Ferdinan, Kazienko, Satapathy, and Cambria}]{yeo2024self}
Wei~Jie Yeo, Teddy Ferdinan, Przemyslaw Kazienko, Ranjan Satapathy, and Erik Cambria. 2024.
\newblock Self-training large language models through knowledge detection.
\newblock \emph{arXiv preprint arXiv:2406.11275}.

\bibitem[{Zhang et~al.(2019)Zhang, Kishore, Wu, Weinberger, and Artzi}]{zhang2019bertscore}
Tianyi Zhang, Varsha Kishore, Felix Wu, Kilian~Q Weinberger, and Yoav Artzi. 2019.
\newblock Bertscore: Evaluating text generation with bert.
\newblock \emph{arXiv preprint arXiv:1904.09675}.

\bibitem[{Zhou et~al.(2021)Zhou, Ma, and Liu}]{zhou2021trade}
Zhihan Zhou, Liqian Ma, and Han Liu. 2021.
\newblock \href {https://arxiv.org/abs/2105.12825} {Trade the event: Corporate events detection for news-based event-driven trading}.
\newblock \emph{Preprint}, arXiv:2105.12825.

\bibitem[{Zhu et~al.(2021)Zhu, Lei, Huang, Wang, Zhang, Lv, Feng, and Chua}]{zhu2021tat}
Fengbin Zhu, Wenqiang Lei, Youcheng Huang, Chao Wang, Shuo Zhang, Jiancheng Lv, Fuli Feng, and Tat-Seng Chua. 2021.
\newblock Tat-qa: A question answering benchmark on a hybrid of tabular and textual content in finance.
\newblock \emph{arXiv preprint arXiv:2105.07624}.

\bibitem[{Zou et~al.(2023)Zou, Shi, Chen, Deng, Lei, Zeng, Yang, Tong, Xiao, and Zhou}]{zou2023esgreveal}
Yi~Zou, Mengying Shi, Zhongjie Chen, Zhu Deng, ZongXiong Lei, Zihan Zeng, Shiming Yang, HongXiang Tong, Lei Xiao, and Wenwen Zhou. 2023.
\newblock Esgreveal: An llm-based approach for extracting structured data from esg reports.
\newblock \emph{arXiv preprint arXiv:2312.17264}.

\end{thebibliography}

\clearpage
\appendix

\section{Prompt Template}
\label{appendix:prompt}
We use alpaca \citep{alpaca} prompt template to train and test our models as following. 
\begin{user_example}[frametitle={Alpaca Prompt Template}]
{
    \fontsize{10pt}{9pt}\selectfont Below is an instruction that describes a task, paired with an input that provides further context.
    Write a response that appropriately completes the request.\\\\
    \emph{<begin of special token>}\\
    \textbf{\#\#\# Instruction:}\\<Instruction>\\\\
    \textbf{\#\#\# Input:}\\<Input>\\\\
    \emph{<end of special token>}\\
    \textbf{\#\#\# Response:}\\<Output>
}  
\end{user_example}
For the Llama3 model, the special tokens are \emph{<|begin\_of\_text|>} and \emph{<|eot\_id|>}, while for the Mistral model, they are \emph{[INST]} and \emph{[/INST]}.
\section{Financial Tasks Definition and Examples}
\label{appendix:2}
This appendix outlines the definitions of various financial tasks, along with corresponding examples within the financial and ESG domains, as addressed by \textsc{SUSGEN-GPT}. These tasks include Headline Classification (HC), Named Entity Recognition (NER), Relation Extraction (RE), Sentiment Analysis (SA), Financial Question Answering (FIN-QA), Financial Table Question Answering (FIN-TQA), Text Summarization (SUM), and Environmental, Social, and Governance Question Answering (ESG-QA). Each task is designed to tackle specific challenges, providing detailed analysis and insights within the financial and ESG contexts.

\subsection{Headline Classification (HC)}
Headline Classification refers to the process of categorizing news headlines or report titles into predefined classes, aiming to systematically group them based on their content, context, or subject matter for easier identification and analysis.
\linebreak
\linebreak

\begin{user_example}[frametitle={Headline Classification}]
{
    \fontsize{10pt}{9pt}\selectfont \textbf{Instruction:} \\In this task, you're working with English headlines from the MULTIFIN dataset. This dataset is made up of real-world article headlines from a large accounting firm's websites. Your objective is to categorize each headline according to its primary topic. The potential categories are 'Finance', 'Technology', 'Tax \& Accounting', 'Business \& Management', 'Government \& Controls', and 'Industry'. Your response should only include the category that best fits the headline. \\
    \textbf{Input:} \\PwC and Alpina Publisher have announced publication of the book The Critical Few\\
    \textbf{Output:} \\Industry.
}  
\end{user_example}

\subsection{Named Entity Recognition (NER)}
Named Entity Recognition is the task of identifying and classifying proper nouns, such as names of individuals, organizations, locations, and financial entities, within a given text.
\begin{user_example}[frametitle={Named Entity Recognition}]
{
    \fontsize{10pt}{9pt}\selectfont \textbf{Instruction:} \\In the sentences extracted from financial agreements in U.S. SEC filings, identify the named entities that represent a person ('PER'), an organization ('ORG'), or a location ('LOC'). The required answer format is: 'entity name, entity type'. \\
    \textbf{Input:} \\SUBORDINATED LOAN AGREEMENT HERBERT SMITH LLP Page 1 of 12 7 - December 2007 TABLE OF CONTENTS Clause Headings Page 1.\\
    \textbf{Output:} \\HERBERT SMITH, PER
}  
\end{user_example}

\subsection{Relation Extraction (RE)}
Relation Extraction involves detecting relationships between entities in text, such as acquisitions, partnerships, regulatory actions, or other interactions.
\begin{user_example}[frametitle={Relation Extraction}]
{
    \fontsize{10pt}{9pt}\selectfont \textbf{Instruction:} \\Given the input sentence, please extract the subject and object containing a certain relation in the sentence according to the following relation types, in the format of \"relation1: word1, word2; relation2: word3, word4\". Relations include: product/material produced; manufacturer; distributed by; industry; position held; original broadcaster; owned by; founded by; distribution format; headquarters location; stock exchange; currency; parent organization; chief executive officer; director/manager; owner of; operator; member of; employer; chairperson; platform; subsidiary; legal form; publisher; developer; brand; business division; location of formation; creator. \\
    \textbf{Input:} \\While chief executive officer Tim Cook has succeeded in introducing an entirely new category with the Apple Watch, sales remain modest, indicating that Apple will have to keep relying on the iPhone to fuel growth.\\
    \textbf{Output:} \\position\_held: Tim Cook, chief executive officer
}  
\end{user_example}

\subsection{Sentiment Analysis (SA)}
Sentiment Analysis is the task of evaluating the emotional tone in a text to determine whether it conveys a positive, negative, or neutral sentiment.
\begin{user_example}[frametitle={Sentiment Analysis}]
{
    \fontsize{10pt}{9pt}\selectfont \textbf{Instruction:} \\
    What sentiment is expressed in this statement? Choose from Positive, Negative, or Neutral. \\
    \textbf{Input:} \\
    Amidst high expectations, Qualcomm's first-quarter projection fell short of analyst estimates, indicating a potential dip in profitability. The company's estimated profit range of 46 to 50 cents per share, excluding certain items, contrasts sharply with the analyst consensus of 61 cents per share, signaling a possible negative market response. \\
    \textbf{Output:} \\
    Neutral
}  
\end{user_example}

\subsection{Financial Question Answering (FIN-QA)}
Financial Question Answering is the task of providing accurate responses to queries based on financial data, reports, and documents, helping users understand complex financial information and make informed decisions.
\begin{user_example}[frametitle={Financial Question Answering}]
{
    \fontsize{10pt}{9pt}\selectfont \textbf{Instruction:} \\
    Read the text below and answer the questions by performing the appropriate numeric operation. \\
    \textbf{Input:} \\
    The NASDAQ index climbed up to the beginning of the opening, but the government bond yields reached a level of nearly 2\%.\\KOSDAQ is trading at 841.36, down 11.47 points (1.34\%) from the previous trading day.Individuals and institutions bought 80 billion won and 18.5 billion won, respectively.On the other hand, foreigners sell 92.8 billion won.\\Question: Where did you buy more individuals and institutions in the KOSDAQ market? \\
    \textbf{Output:} \\
    In the KOSDAQ market, more individuals and institutions are bought.The individual bought 80 billion won, while the agency bought 18.5 billion won.Therefore, individuals bought more than institutions.Therefore, the answer is an individual.
}  
\end{user_example}

\subsection{Financial Table Question Answering (FIN-TQA)}
Financial Table Question Answering extends traditional question answering by focusing on structured data, specifically extracting and interpreting information from financial tables.
\begin{user_example}[frametitle={Financial Table Question Answering}]
{
    \fontsize{10pt}{9pt}\selectfont \textbf{Instruction:} \\
    Please provide your answer to the financial question based on the context. \\
    \textbf{Input:} \\
    Entergy Corporation and Subsidiaries Management 2019’s Financial Discussion and Analysis: A result of the Entergy Louisiana and Entergy Gulf States Louisiana business combination, results of operations for 2015 also include two items that occurred in October 2015: 1) a deferred tax asset and resulting net increase in tax basis of approximately \$334 million, and 2) a regulatory liability of \$107 million (\$66 million net-of-tax) as a result of customer credits to be realized by electric customers of Entergy Louisiana, consistent with the terms of the stipulated settlement in the business combination proceeding. See Note 2 to the financial statements for further discussion of the business combination and customer credits......
    Net Revenue Utility: Following is an analysis of the change in net revenue comparing 2015 to 2014.
    \begin{center}
    \begin{tabular}{|p{0.5\linewidth}|r|}
    \hline
    \textbf{Amount (in millions)} & \\
    \hline
    2014 Net Revenue & \$5735 \\
    \hline
    Retail Electric Price & 187 \\
    \hline
    Volume/Weather & 95 \\
    \hline
    Waterford 3 Replacement Steam Generator Provision & -32 (32) \\
    \hline
    MISO Deferral & -35 (35) \\
    \hline
    Louisiana Business Combination Customer Credits & -107 (107) \\
    \hline
    Other & -14 (14) \\
    \hline
    2015 Net Revenue & \$5829 \\
    \hline
    \end{tabular}
    \end{center}
    
    The retail electric price variance is primarily due to: 2022 Formula Rate Plan increases at Entergy Louisiana, as approved by the LPSC, effective December 2014 and January 2015; 2022 an increase in Energy Efficiency Rider revenue primarily due to increases in the Energy Efficiency Rider at Entergy Arkansas, as approved by the APSC, effective July 2015 and July 2014, and new Energy Efficiency Riders at Entergy Louisiana and Entergy Mississippi that began in the fourth quarter of 2014; and 2022 an annual net rate increase at Entergy Mississippi of \$16 million, effective February 2015, as a result of the MPSC order in the June 2014 rate case. See Note 2 to the financial statements for a discussion of rate and regulatory proceedings.
    
    Question: What is the net change in net revenue during 2015 for Entergy Corporation?
    
    \textbf{Output:} \
    94.0
}  
\end{user_example}
\subsection{Text Summarization}
Text Summarization involves generating concise and coherent summaries from longer documents, focusing on extracting the most relevant information while preserving the overall meaning.
\begin{user_example}[frametitle={Text Summarization}]
{
    \fontsize{10pt}{9pt}\selectfont \textbf{Instruction:} \\
    You are given a text consists of multiple sentences. Your task is to perform abstractive summarization on this text. Use your understanding of the content to express the main ideas and crucial details in a shorter, coherent, and natural sounding text. \\
    \textbf{Input:} \\
    BEIJING, April 9, 2020 /PRNewswire/ -- Baidu, Inc. (NASDAQ: BIDU) (\"Baidu\" or the \"Company\"), a leading search engine, knowledge and information centered Internet platform and AI company, today announced that pursuant to directives of relevant PRC regulators, Baidu has suspended updating its content on certain newsfeeds channels within Baidu App and conduct maintenance, beginning from April 8, 2020. The Company expects that the suspension may have impact on the marketing services revenue related to the suspended channels. The Company will undertake additional measures to fully comply with the directives of the regulators, ... Chinese governmental policies relating to the Internet and Internet search providers, and general economic conditions in China and elsewhere. ... SOURCE Baidu, Inc. \\
    \textbf{Output:} \\
    Baidu Takes Measures to Comply with Government Directives
}  
\end{user_example}

\subsection{Environmental, Social, and Governance Question Answering (ESG-QA)}
Environmental, Social, and Governance (ESG) Question Answering is a specialized task aimed at answering questions related to sustainability, corporate responsibility, and governance practices.
\begin{user_example}[frametitle={ESG Question Answering}]
{
    \fontsize{10pt}{9pt}\selectfont \textbf{Instruction:} \\
    As an ESG specialist at NASDAQ, you help companies improve their ESG scores and can answer any ESG-related questions. \\
    \textbf{Input:} \\
    What are the necessary steps to implement free trade  policies in my company? \\
    \textbf{Output:} \\
    Implementing free trade policies in a company involves several necessary steps. Here is a general outline of the process:\\1. Research and Evaluate: Start by researching the concept of free trade understanding its impact benefits and challenges. Evaluate how adopting free trade policies will align with your company's goals vision and industry dynamics.\\2. Assess Current Trade Policies: Evaluate your current trade policies and identify any barriers or restrictions that could hinder free trade. Analyze your existing agreements tariff structures trade regulations and any protectionist measures.\\
    ...\\
    11. Monitor and Review: Continuously monitor the impact of your free trade policies evaluate their effectiveness and make necessary adjustments. Stay updated on evolving trade regulations market trends and any political developments that might affect international trade.\\
    Remember that specific steps may vary depending on your industry company size and market conditions. Consulting with trade experts and professionals in your specific field can provide the necessary guidance tailored to your unique circumstances.
}  
\end{user_example}

These examples illustrate the diverse capabilities of \textsc{SUSGEN-GPT} in processing and analyzing financial data.

\section{The Composition of \textsc{SusGen-30k}}
\label{appendix:1}
\subsection{Data Statistics}
\label{appendix:1-data}
This section details the various components and sources of data included in the SUSGEN dataset, providing an overview of the dataset's structure and content.

Each dataset listed in Table \ref{tab:dataset} is meticulously curated and split to optimize model training and evaluation, covering a diverse range of scenarios and applications in financial NLP. This approach ensures that the model performs well across these specific tasks and integrates seamlessly with real-world financial applications, providing robust and reliable insights.

\begin{figure*}[ht]
    \centering
    \includegraphics[width=\linewidth]{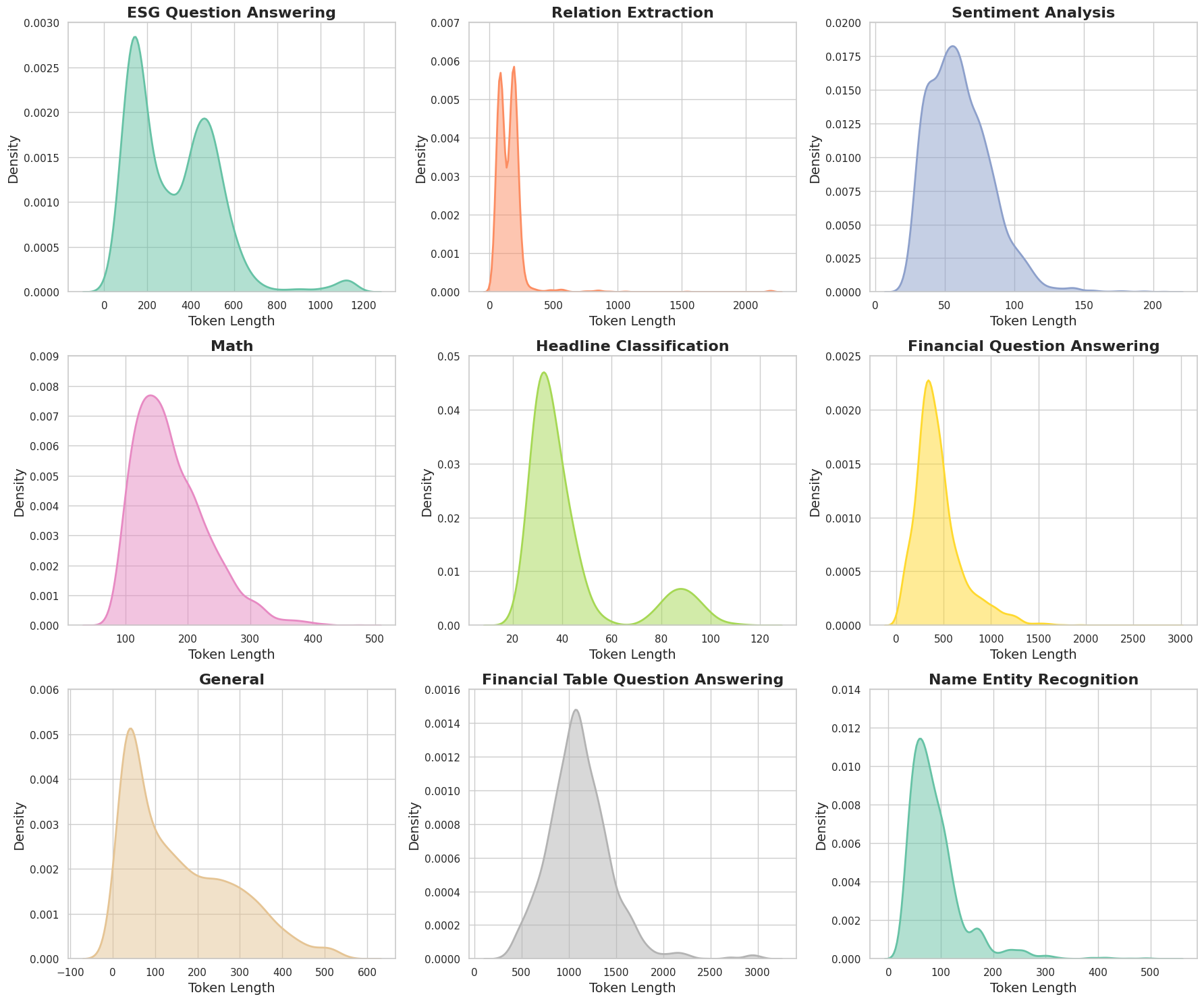}
    \caption{Token length distribution across the 9 sections of \textsc{SusGen-30k} dataset.}
    \label{fig:susgen30k_dist}
\end{figure*}

We also provide the token length distribution regarding individual task in our \textsc{SusGen-30k} dataset, as shown in Figure \ref{fig:susgen30k_dist}.

\subsection{LLMs automatic pre-processing pipeline}
This section explains how we construct the automatic pre-processing pipeline utilizing large language models (LLMs) to handle our aggregated data, which includes five steps in total.
\subsubsection{Translation}
The first step in our pre-processing pipeline is handling multilingual data, as the aggregated dataset contains content in several languages, including German, French, and Korean, etc. We begin by detecting these non-English portions of the data. Once identified, LLMs are used to automatically translate the non-English text into English. This translation process ensures that all data, regardless of its original language, is standardized in English, which is essential for consistent downstream processing in subsequent steps.
\subsubsection{Reformatting}
In the second step of our pipeline, we prepare the data for supervised fine-tuning LLMs. To achieve this, all data is reformatted into an instruction-following format, similar to the prompt templates shown in Appendix \ref{appendix:prompt}. This involves structuring each data point with clear sections for instruction, input, and output. By converting the data into this format, we ensure that it aligns with the instruction-based learning paradigm \citep{alpaca,instructgpt}, optimizing it for fine-tuning large models to follow and execute tasks as instructed.
\subsubsection{Anonymization}
As illustrated in Figure \ref{fig:data_construction}, our dataset is composed of two key parts, one of which is sourced from TCFD Hub. This portion of the data was extracted from publicly available PDFs and reformatted into an instruction-following format. However, the output sections of this data often contain a significant amount of company-specific information. Although these reports are publicly accessible, we aim to minimize potential model biases and protect data privacy. To achieve this, we employ LLMs to anonymize the entity-related information, ensuring that all company names and identifying details are removed. This step helps safeguard privacy while maintaining the integrity of the dataset. We use Mixtral-8x22B-Instruct-v0.1\footnote{\url{https://huggingface.co/mistralai/Mixtral-8x7B-v0.1}} for this step using the following prompt to process those output:
\begin{user_example}[frametitle={Prompt for Anonymization}]
{
    \fontsize{10pt}{9pt}\selectfont [INST] \\Process the text following the instructions below:\\
    1.Replace all the specific company entity name with "we" or "our company".\\
    2.Replace other private information with generic terms.\\
    Text: \{Text\} \\ {[/INST]}
}  
\end{user_example}
\subsubsection{Augmentation}
After processing the collected data, we noticed that many datasets contain fixed instruction formats. For example, in Sentiment Analysis, the instruction might always be something like: “Please analyze the sentiment in the input and answer ‘negative,’ ‘positive,’ or ‘neutral’.” This lack of variation in instructions can lead to over-fitting during model training, as the model may become overly accustomed to fixed prompts. To address this issue, we augment the instruction sections by introducing diverse prompts. By generating varied instructions, we inject noise and variation into the data, reducing the likelihood of the model over-fitting to a single fixed instruction format. This augmentation step ensures the model is exposed to a broader range of instructions, improving its generalization capabilities. The prompt that we used is shown below.
\begin{user_example}[frametitle={Prompt for Augmentation}]
{
    \fontsize{10pt}{9pt}\selectfont [INST] \\Process the text following the instructions below:\\
    1.Rephrase the whole text without change original meaning and elements.\\
    2.Adjust the processed text to similar length as the original text.\\
    3.Ensure the text is coherent and fluent and output the final text.\\
    Text: \{Text\} \\ {[/INST]}
}  
\end{user_example}
\subsubsection{Synthesizing}
For certain tasks, such as ESG-QA, the dataset contains fewer samples compared to other tasks, creating a challenge for effective model training. Drawing inspiration from Common-7B \citep{li2024common}, which demonstrated that training LLMs on synthetic data can improve performance in specific domains, we applied a similar approach to augment the data for these low-sample tasks. The data synthesis process follows a similar methodology to the third (Anonymization) and fourth (Augmentation) steps, but with some key differences. We increased the temperature in the generation process to produce more diverse outputs, and we applied augmentation to both the instruction and output sections. Finally, as with the anonymization step, we ensured that all synthesized outputs were anonymized to protect privacy. This approach resulted in an expanded dataset for the tasks with fewer data samples, enhancing model performance in those areas while maintaining data integrity.

\begin{table*}[ht]
    \centering
    \resizebox{\linewidth}{!}{
    \begin{tabular}{@{} l l r c r r r @{}}
    \toprule
    Task  & Dataset & Train & Language & Test  & Final & Comment \\
    \midrule
    General & Alpaca-52k \citep{alpaca} & 52,000 & EN & \ding{55} & 3,000 & \\ 
    Arithmetic & GSM-8k \citep{cobbe2021gsm8k} & 7,473 & EN & 1,319 & 3,000 & \\ 
    \midrule
    HC & fingpt-headline-cls \citep{wang2023fingpt} &  82,200 & EN & 20,500 &  1,500 & CLS \\ 
    HC & fingpt-headline \citep{wang2023fingpt} &  82,200 & EN & 20,500 &  1,500 & Instr Diff \\ 
    HC & FLUE-headline \citep{shah2022flue} &  80,000 & EN & \ding{55} &  0 & \ding{55} \\ 
    HC & flare-multifin-en \citep{xie2024finben} &  \ding{55} & EN & 546 &  500 & CLS \\ 
    HC & flare-mlesg-en \citep{xie2024finben} &  \ding{55} & EN & 300 &  300 & ESG-CLS \\
    \midrule
    NER & fingpt-ner-cls \citep{wang2023fingpt} & 13,500 & EN & 3,500 &  2,700 & CLS \\ 
    NER & fingpt-ner \citep{wang2023fingpt} & 511 & EN & 98 &  500 & \\
    NER & flare-ner \citep{xie2024finben} & 408 & EN & 98 & 300 & valid103\\
    NER & flare-finer-ord \citep{xie2024finben} & \ding{55} & EN & 1,075 & 1,075 & \\
    \midrule
    RE & fingpt-finred \citep{wang2023fingpt} &  27,600 & EN &  5,112 & 5,112 & RE+CLS \\
    RE & fingpt-finred-re \citep{wang2023fingpt} &  11,400 & EN &  2,140 &  1,750 & RE \\
    RE & fingpt-finred-cls \citep{wang2023fingpt} &  48,500 & EN &  8,930 &  1,750 & CLS \\
    RE & flare-finarg-ecc-auc-test \citep{xie2024finben} & \ding{55} & EN &  969 & 0 & RE+CLS \\
    RE & flare-causal20-sc-test \citep{xie2024finben} & \ding{55} & EN &  8,628 & 8,628 & RE+CLS \\
    \midrule
    SA & esg-sentiment & 611 & EN & 93 &  843 & ESG\\ 
    SA & enhanced-financial-phrasebank & 4,850 & EN   &  \ding{55} & 1,457 \\
    SA & FIN\_NUMBER-SA/train (\href{https://huggingface.co/FINNUMBER}{link}) & 4,680 & KO & \ding{55} & xx & ESG \\
    SA & fingpt-sentiment \citep{wang2023fingpt} & 76,800 & EN & \ding{55} & 800 &  \\
    SA & fingpt-sentiment-cls \citep{wang2023fingpt} & 47,600 & EN & \ding{55} & 400 & CLS  \\    
    SA & FLUE-sentiment \citep{shah2022flue} & 4850 & EN   &  \ding{55} & 0 & \ding{55} \\ 
    SA & flare-fiqasa \citep{xie2024finben} & 750 & EN   &  235 & 235 & valid188\\ 
    SA & flare-fomc \citep{xie2024finben} & \ding{55} & EN   &  496 & 496 & valid188\\
    \midrule
    FIN-QA & FIN\_NUMBER-EQA/train                & 400 & KO   &  \ding{55} &  400 \\
    FIN-QA  & FIN\_NUMBER-BQA/train                 &      400 & KO   &  \ding{55} &  400 & CLS\\
    FIN-QA & FIN\_NUMBER-MCQA/train                 &      400 & KO   &  \ding{55} &  398 & CLS\\
    FIN-QA & FIN\_NUMBER-NQA-ARI/train &      400 & KO   &  \ding{55} &  398 \\
    FIN-QA & FIN\_NUMBER-NQA-COM/train                &      400 & KO   &  \ding{55} &  399 \\
    FIN-QA & FIN\_NUMBER-NQA-EXT/train                &  400 & KO   &  \ding{55} &  397 \\
    FIN-QA & flare-cfa/test \citep{xie2024finben} & \ding{55} & EN   &  1030 & 0 & CLS\\
    FIN-QA & fingpt-fiqa\_qa \citep{wang2023fingpt} &  17,100 & EN &  \ding{55} &  708 \\
    FIN-QA  & fingpt-fineval \citep{wang2023fingpt} & 1,060 & ZH & 265 & 0 & CLS \\
    FIN-QA & flare-finqa  \citep{xie2024finben}              & 6250 & EN & 1147 & 400 \\
    FIN-QA & flare-fsrl  \citep{xie2024finben}             & \ding{55} & EN & 97 & 97 & \\
    \midrule
    FIN-TQA  & fingpt-convfinqa \citep{wang2023fingpt} & 11,100 & EN & 1,490 & 1,000 & \\
    FIN-TQA  & flare-convfinqa \citep{xie2024finben} & 8890 & EN & 1,490 & 2,500 &  \\
    FIN-TQA  & flare-tatqa \citep{xie2024finben} & \ding{55} & EN & 1,668 & 1,668 &  \\
    \midrule
    SUM  & flare-edtsum-test \citep{xie2024finben} & \ding{55} & EN & 2000 & 2000 &  \\
    \midrule
    ESG-QA  & ESG-Chat & 914 & EN   &  \ding{55} &  914 \\ 
    ESG-QA  & TCFD\_QA & 260 & EN   &  \ding{55} & 1669 \\ 
    ESG-QA & salmasally & 417 & FR   &  \ding{55} & 417 \\ 
    \bottomrule
    \end{tabular}
    }
    \caption{
      \textbf{Composition of our \textsc{SusGen-30k} dataset.} 
      We report the list of datasets and associated splits used to build the dataset. We mainly focus on eight following tasks in the datasets in order to let the model cover most applications in the financial NLP domain. HC: Headline Classification. NER: Named Entity Recognition. RE: Relation Extraction. SA: Sentiment Analysis.  FIN-QA: Financial Question Answering. FIN-TQA: Financial Table Question Answering. SUM: Text Summary. ESG-QA: Environmental, Social, and Governance Question Answering. Additionally, we integrate portions of Alpaca and GSM8K into our dataset to mitigate the risk of model over-fitting.  
      }
    \label{tab:dataset}
\end{table*}

\section{Our Ablation Study Results about Training Data Scale}
\label{appendix:3}
\subsection{Quantitative Ablation Study}
This section showcases the quantitative results of our ablation studies, focusing on the training data scaling effect on our models' performance.

Table \ref{tab:datascale} presents the zero-shot and few-shot performance of various LLMs, including different configurations of \textsc{SusGen-GPT} with different training dataset sizes of 10k, 20k, and 30k respectively. The metrics evaluated include F1 score, Micro F1, Entity F1, Exact Match Accuracy, Rouge, and BertScore across 8 tasks across multiple financial nlp benchmarks. The results indicate that increasing the dataset size significantly improves the performance of SusGen-GPT, especially after supervised fine-tuning. 

\begin{table*}[ht]
		\centering
		\scriptsize
 \renewcommand{\arraystretch}{1.2}
 \scalebox{0.8}{
	\begin{tabular}{ll|cccc|cccc|cccc}
			\toprule
			\textbf{Datasets} 
                &\textbf{Metrics}
            &\makecell{\textbf{SusGen}\\\textbf{GPT-10k}\\\textbf{Llama3}\\\textbf{8B}\\\textbf{Instruct}}
            &\makecell{\textbf{SusGen}\\\textbf{GPT-10k}\\\textbf{Llama3}\\\textbf{8B}}
            &\makecell{\textbf{SusGen}\\\textbf{GPT-10k}\\\textbf{Mistral}\\\textbf{7B}\\\textbf{Instruct}\\\textbf{v0.3}}
            &\makecell{\textbf{SusGen}\\\textbf{GPT-10k}\\\textbf{Mistral}\\\textbf{7B}\\\textbf{v0.3}}
            &\makecell{\textbf{SusGen}\\\textbf{GPT-20k}\\\textbf{Llama3}\\\textbf{8B}\\\textbf{Instruct}}
            &\makecell{\textbf{SusGen}\\\textbf{GPT-20k}\\\textbf{Llama3}\\\textbf{8B}}
            &\makecell{\textbf{SusGen}\\\textbf{GPT-20k}\\\textbf{Mistral}\\\textbf{7B}\\\textbf{Instruct}\\\textbf{v0.3}}
            &\makecell{\textbf{SusGen}\\\textbf{GPT-20k}\\\textbf{Mistral}\\\textbf{7B}\\\textbf{v0.3}}
            &\makecell{\textbf{SusGen}\\\textbf{GPT-30k}\\\textbf{Llama3}\\\textbf{8B}\\\textbf{Instruct}}
            &\makecell{\textbf{SusGen}\\\textbf{GPT-30k}\\\textbf{Llama3}\\\textbf{8B}}
            &\makecell{\textbf{SusGen}\\\textbf{GPT-30k}\\\textbf{Mistral}\\\textbf{7B}\\\textbf{Instruct}\\\textbf{v0.3}}
            &\makecell{\textbf{SusGen}\\\textbf{GPT-30k}\\\textbf{Mistral}\\\textbf{7B}\\\textbf{v0.3}}

			\\\midrule
   
            \textcolor{black}{FiQASA}
			&F1&0.63&0.50&0.63& 0.60 &0.66&0.56&0.64&0.76&0.66&0.46&\textcolor{black}{\textbf{0.72}}&0.63\\

             \textcolor{black}{FOMC}
			&F1&\textcolor{black}{0.61}&0.37&\textcolor{black}{0.59}&0.46&0.63&0.57 & 0.60&0.60&\textcolor{black}{\textbf{0.70}}&0.64&0.69&0.67\\\hline
            \textcolor{black}{MultiFin}&MicroF1&0.30&0.00&0.40&0.39&0.43&0.42&0.51&0.50&0.46&0.43&\textcolor{black}{\textbf{0.52}}&0.51\\

            \textcolor{black}{MLESG}
			&MicroF1&0.30\textcolor{black}&0.03&0.41&0.06&0.25&0.08&0.44&0.10&0.28&0.04&\textcolor{black}{\textbf{0.51}}&0.14\\\hline

\textcolor{black}{NER}
			&EntityF1&0.27&0.10&0.02&0.17&0.21&0.25&0.25&0.24&\textbf{0.35}&0.33&0.25&0.31\\

            \textcolor{black}{FINER-ORD}
			&EntityF1&0.08&0.00&0.14&0.11&0.09&0.09&0.14&0.11&0.11&0.10&\textbf{0.18}&0.17\\\hline
    
 \textcolor{black}{FinRED}
			&F1&\textcolor{black}{0.04}&0.02&\textcolor{black}{0.06}&0.05&0.16&0.21&0.11&0.09&0.16&\textbf{0.23}&0.19&0.17\\

            \textcolor{black}{SC}
			&F1&\textcolor{black}{0.67}&0.23&\textcolor{black}{0.85}&0.85&0.72&0.48&0.86&0.84&0.86&0.41&\textbf{0.96}&\textbf{0.96}\\\hline
			FinQA
			&EmAcc&0.35&0.46&0.33&0.37&0.37&0.51&0.53&0.54&0.38&0.54&\textbf{0.57}&\textbf{0.57}\\\hline
			 TATQA
			&EmAcc&0.59&{0.41}&\textcolor{black}{0.67}&0.58&0.59&0.57&0.69&0.65&0.69&0.62&\textbf{0.80}&0.65\\
			ConvFinQA
			&EmAcc&0.58&0.35&0.54&0.43&0.59&0.60&0.70&0.64&0.70&\textbf{0.73}&0.69&0.58\\\hline
			\multirow{2}{*}{EDTSUM}
			&Rouge-1&\textcolor{black}{0.10}&\textcolor{black}{0.12}&\textcolor{black}{0.14}&0.10&0.20&0.14&0.20&0.19&0.21&0.22&\textbf{0.27}&0.23\\
			&Rouge-2&\textcolor{black}{0.08}&\textcolor{black}{0.03}&\textcolor{black}{0.10}&0.04&0.08&0.04&0.11&0.10&0.08&0.05&\textbf{0.13}&0.11\\
			&Rouge-L&\textcolor{black}{0.10}&\textcolor{black}{0.10}&\textcolor{black}{0.19}&0.08&0.16&0.15&\textbf{0.22}&0.10&0.16&0.19&\textbf{0.22}&0.15\\
			&BertScore&\textcolor{black}{0.44}&0.47&\textcolor{black}{0.43}&0.45&0.50&0.49&0.45&0.50&0.51&0.48&\textcolor{black}{\textbf{0.54}}&0.51\\\hline
            
			\multirow{2}{*}{TCFD-BENCH}
			&Rouge-L&0.13&\textcolor{black}{0.10}&0.17&0.13&0.14&0.11&0.18&0.16&0.14&0.12&\textbf{0.20}&0.15\\
			&BertScore&\textcolor{black}{0.22}&\textcolor{black}{0.28}&\textcolor{black}{0.25}&0.20&0.31&0.29&\textbf{0.40}&0.36&0.31&0.29&\textbf{0.40}&0.37\\
			&METEOR&\textcolor{black}{0.15}&\textcolor{black}{0.15}&\textcolor{black}{0.14}&0.15&0.20&0.19&0.22&0.21&0.24&0.21&\textbf{0.27}&0.23\\
			&BLEU-1&\textcolor{black}{0.12}&\textcolor{black}{0.11}&\textcolor{black}{0.17}&0.18&0.20&0.19&0.27&0.20&0.20&0.17&\textbf{0.32}&0.19\\
			&BLEU-2&\textcolor{black}{0.05}&\textcolor{black}{0.03}&\textcolor{black}{0.05}&0.05&0.05&0.03&\textbf{0.12}&0.09&0.05&0.03&\textbf{0.12}&0.10\\
			&BLEU-3&\textcolor{black}{0.02}&\textcolor{black}{0.01}&\textcolor{black}{0.03}&0.02&0.02&0.01&0.06&0.04&0.02&0.01&\textbf{0.07}&0.04\\
			&BLEU-4&0.01&0.00&0.02&0.01&0.01&0.01&0.04&0.02&0.01&0.01&\textbf{0.05}&0.03\\
			\bottomrule
        \end{tabular}}
		\caption{Ablation study results of our models on 8 different financial NLP tasks, using three different subset datasets of 10k, 20k, and 30k examples. The suffix in each model name denotes the base model used for initialization.}	
  	\label{tab:datascale}
	\end{table*} 

\begin{table*}[t]
\small
\centering
\resizebox{\textwidth}{!}{
\begin{tabular}{>{\centering\arraybackslash}m{0.1\textwidth}m{0.7\textwidth}m{0.2\textwidth}}
\hline
Models & Example Prompt & Example Output \\ \hline


Mistral7B-Instruct-v0.3 & \textit{\textbf{Question Answering:}} \textcolor{forestgreen}{TCFD stands for Task Force on Climate-related Financial Disclosures. It's an initiative by the Financial Stability Board (FSB) to develop a framework for companies and other organizations to disclose climate-related financial risks and opportunities.} \textcolor{goldenrod}{The goal of this framework is to help investors, lenders, and insurance underwriters assess and price these risks more accurately, thereby promoting more informed decision making and better management of climate risk.} \textcolor{blue}{The TCFD format provides guidance on how organizations can report their exposure to climate-related risks and opportunities, including physical risks such as extreme weather events or sea level rise, transition risks related to policy changes or technological shifts, and business opportunities arising from low carbon technologies or new markets.} By adopting the TCFD format, organizations can improve transparency and comparability of climate-related information, which can ultimately lead to more effective management of climate risk and more sustainable investment decisions. & \textcolor{forestgreen}{Basic understanding of TCFD, } \textcolor{goldenrod}{but suboptimal text length and unclear structure.} \\ \hline

\textsc{SusGen-GPT}-10k-Mistral7B-Instruct-v0.3 & \textit{\textbf{Question Answering:}} \textcolor{forestgreen}{TCFD stands for Task Force on Climate-related Financial Disclosures. It's an initiative by the Financial Stability Board (FSB) to develop a framework for companies and organizations to disclose information related to climate-related risks and opportunities,} \textcolor{goldenrod}{with the goal of improving transparency and understanding of these factors for investors, lenders, insurers, and other stakeholders.} \textcolor{blue}{The TCFD recommendations cover four areas: Governance, Strategy, Risk Management, Metrics and Targets. Companies that adopt TCFD reporting provide detailed information about their approach to managing climate risk and opportunity, including their governance structures, strategy for addressing climate issues, and metrics used to measure progress towards sustainability goals.} This information helps stakeholders assess the resilience of an organization's business model and its ability to manage climate-related risks effectively. By providing clear, consistent, and comparable information, TCFD aims to enhance market efficiency, reduce financial risks, and support informed decision-making. In summary, TCFD is a set of recommendations aimed at enhancing an organization's transparency regarding climate-related risks and opportunities, enabling better investment decisions and improved management of climate-related risks. & \textcolor{forestgreen}{Good understanding of TCFD, } \textcolor{goldenrod}{improved text length}, \textcolor{blue}{but unclear explanation structure.} \\ \hline

\textsc{SusGen-GPT}-20k-Mistral7B-Instruct-v0.3 & \textit{\textbf{Question Answering:}} \textcolor{forestgreen}{TCFD stands for Task Force on Climate-related Financial Disclosures. It's an international organization founded by the Financial Stability Board (FSB) to develop recommendations for companies and other organizations to disclose climate-related risks and opportunities in their financial reporting. } \textcolor{goldenrod}{The TCFD framework includes four core areas of climate-related risk management: Governance, Strategy, Risk Management, Metrics and Targets}. \textcolor{goldenrod}{Companies that adopt TCFD disclosures are expected to provide transparent and consistent information about how they identify, assess, and manage climate-related risks and opportunities, enabling investors, lenders, and other stakeholders to make informed decisions based on this information. } \textcolor{magenta}{By using TCFD format, organizations can demonstrate their commitment to transparency and accountability regarding climate change and its potential impact on their business operations and financial performance}. & \textcolor{forestgreen}{Excellent TCFD understanding, } \textcolor{goldenrod}{clear explanation structure}, \textcolor{magenta}{but poor text length.} \\ \hline

\textsc{SusGen-GPT}-30k-Mistral7B-Instruct-v0.3 & \textit{\textbf{Question Answering:}}  \textcolor{forestgreen}{TCFD stands for Task Force on Climate-related Financial Disclosures. It is an international organization founded by the Financial Stability Board (FSB) in 2015 to develop recommendations for companies and other organizations to disclose climate-related risks and opportunities in a more transparent way. } \textcolor{goldenrod}{The TCFD framework focuses on four key areas of climate-related risk management: Governance, Strategy, Risk Management, Metrics and Targets}. \textcolor{blue}{Companies that adopt TCFD reporting are expected to provide detailed information about their climate-related risks and opportunities, enabling investors and stakeholders to make informed decisions based on this data. } \textcolor{magenta}{TCFD reports are typically presented in a standardized format, allowing for easy comparison between different organizations. This promotes greater transparency and accountability when it comes to addressing climate-related issues within financial portfolios.} \textcolor{blue}{By providing clear and consistent information, TCFD aims to improve the overall understanding of climate-related risks and opportunities, and help organizations better manage these factors to mitigate potential negative impacts and capitalize on emerging opportunities.} In summary, TCFD format refers to the specific disclosure requirements set forth by the TCFD for organizations to report on climate-related risks and opportunities in a transparent and comparable manner. & \textcolor{forestgreen}{Deep understanding of TCFD, } \textcolor{goldenrod}{appropriate text length, } \textcolor{blue}{clear answer structure, } \textcolor{magenta}{and minimal noise}. \\ \hline
\end{tabular}
}
\caption{Ablation study results of our models on open-ended question answering, using three different subset datasets
of 10k, 20k, and 30k examples. The suffix in each model name denotes the base model used for initialization.}
\label{fig:ablation_report_table}
\end{table*}

\subsection{Qualitative Ablation Study}
In this section, we present the qualitative ablation results of our models to test their ability to generalize to open-ended tasks. Since many tasks are close-ended and can be demonstrated through quantitative results, we selected an open-ended case to showcase qualitative performance.

The prompt we give to the model is "What is tcfd format in the context of climate change?". As shown in the Table \ref{fig:ablation_report_table}, revealing clear trends in model performance as dataset size increases from 10k to 30k. As demonstrated in the table, the \textsc{SusGen-GPT}-30k-Mistral7B-Instruct-v0.3 model produces the most coherent, well-structured, and detailed response to the open-ended prompt. It exhibits a deep understanding of the TCFD framework, provides an appropriate text length, and features minimal noise, reflecting its strong ability to handle complex tasks. The \textsc{SusGen-GPT}-20k model also demonstrates a strong grasp of TCFD, but lacks the text length and detail seen in the 30k model. Meanwhile, the 10k model, while offering a solid response, presents a less structured and slightly less clear explanation. Lastly, the untrained or minimally trained models show noticeable gaps in their answers, with shorter, less precise responses. This progression suggests that increasing the training dataset size significantly enhances the model’s capacity to deliver high-quality, open-ended outputs.
\section{Sustainability Report Generation Examples}
\label{appendix:4}
We provide examples of sustainability reports generated by \textsc{SusGen-GPT}, as shown in Figure \ref{tab:prompt_dev} ,demonstrating the model's ability to create comprehensive and accurate ESG reports. SusGen-GPT's structured format offers significant advantages, establishing it as a superior tool for generating TCFD-format ESG reports. These enhancements ensure that reports produced by SusGen-GPT are not only more informative but also more actionable and easier to comprehend. By categorizing risks and opportunities into specific time horizons (short-term, medium-term, and long-term), SusGen-GPT provides a clear roadmap for addressing climate-related challenges and leveraging opportunities. Each identified risk and opportunity is linked to specific actions and goals, making the report practical and aligned with the company's strategic objectives.

Additionally, the improved contextual clarity provided by SusGen-GPT allows users to better understand the implications of each risk and opportunity. This is essential for stakeholders who depend on these reports to make informed decisions. The detailed action points offered by SusGen-GPT assist in developing concrete strategies for sustainability, enhancing operational efficiencies, improving reputation, and ensuring compliance with regulatory requirements.

In contrast, while ChatReport delivers a general overview of climate-related risks and opportunities, it lacks the depth and specificity present in SusGen-GPT’s output. ChatReport's responses are less structured and do not consistently link risks to specific time horizons, making it more challenging for users to prioritize actions and comprehend the timeline for implementation.



\end{document}